\documentclass{article}
\usepackage{jfrExamplee}
\usepackage{graphicx}
\usepackage{apalike}
\usepackage{setspace}
\usepackage{authblk}
\usepackage{subfigure}
\usepackage{cite}
\usepackage{amsmath,amssymb,amsfonts}
\usepackage{algorithmic}
\usepackage{bbding}
\usepackage{bm}
\usepackage{float}
\usepackage{diagbox}
\usepackage{multirow}
\usepackage{harpoon}
\usepackage{stfloats}
\usepackage{float}
\usepackage{booktabs}
\usepackage{caption,setspace}
\usepackage{multirow}
\usepackage{changes}

\newtheorem{theorem}{Theorem}

\newcommand{\tabincell}[2]{\begin{tabular}{@{}#1@{}}#2\end{tabular}}
\newsavebox\CBox
\def\textBF#1{\sbox\CBox{#1}\resizebox{\wd\CBox}{\ht\CBox}{\textbf{#1}}}
\makeatletter
\def \thanks#1{{\protected@xdef\@thanks{\@thanks\protect\footnotetext{#1}}}}
\makeatother

\title{Traversability Analysis for Autonomous Driving in Complex Environment: A LiDAR-based Terrain Modeling Approach}

\author{Hanzhang Xue$^{1,2}$, Hao Fu$^{1}$, Liang Xiao$^{2}$, Yiming Fan$^{1}$, Dawei Zhao$^{2}$, Bin Dai$^{2}$
	\thanks{This work was supported by the National Natural Science Foundation of China under No.61790565 and No.61803380.}
	\thanks{$^{1}$ College of Intelligence Science and Technology, National University of Defense Technology, Changsha, 410073, P.R. China. email: {\tt\small xuehanzhang13@nudt.edu.cn, fuhao@nudt.edu.cn, fanyimingendeavor@126.com}}%
	\thanks{$^{2}$ Unmanned Systems Technology Research Center, Defense Innovation Institute, Beijing, 100071, P.R. China. email:
	{\tt\small xiaoliang@nudt.edu.cn, adamzdw@163.com, ibindai@163.com}}%
}

\begin{document}

\maketitle

\begin{abstract}
For autonomous driving, traversability analysis is one of the most basic and essential tasks. In this paper, we propose a novel LiDAR-based terrain modeling approach, which could output stable, complete and accurate terrain models and traversability analysis results. As terrain is an inherent property of the environment that does not change with different view angles, our approach adopts a multi-frame information fusion strategy for terrain modeling. Specifically, a normal distributions transform mapping approach is adopted to accurately model the terrain by fusing information from consecutive LiDAR frames. Then the spatial-temporal Bayesian generalized kernel inference and bilateral filtering are utilized to promote the stability and completeness of the results while simultaneously retaining the sharp terrain edges. Based on the terrain modeling results, the traversability of each region is obtained by performing geometric connectivity analysis between neighboring terrain regions. Experimental results show that the proposed method could run in real-time and outperforms state-of-the-art approaches.
\end{abstract}

\section{INTRODUCTION}
\label{Section: Intro}
For autonomous driving, determining the traversable regions in the environment is one of the most essential tasks, which directly affects the safety of unmanned ground vehicles (UGVs). Closely related tasks including road detection \cite{Mei}, ground segmentation \cite{Hamandi}, drivable region detection \cite{Lyu}, etc.

\begin{figure}[t]
	\centering
	\includegraphics[scale=0.6]{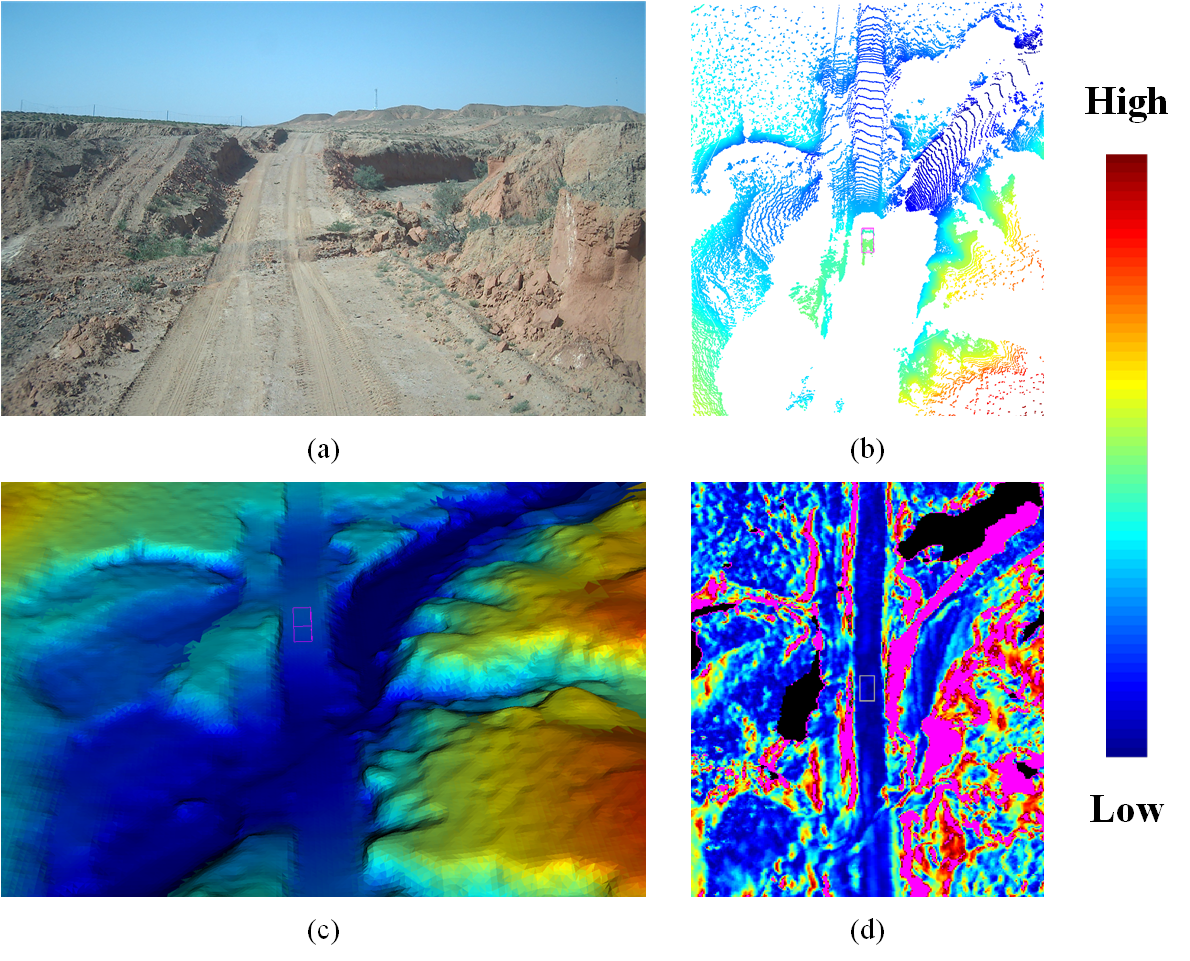} \\
	\caption{(a) is an image of the rugged off-road environment, and (b) displays the single point cloud generated by LiDAR in this environment. The terrain model and cost map generated by the proposed method are shown in (c) and (d), respectively ((b) and (c) are colored by terrain elevation, and (d) is colored by travel cost. Non-traversable and unreachable regions in (d) are colored by pink). }
	\label{fig_illustration}
\end{figure}

By analyzing these related works, we found that most existing approaches treat traversability analysis as a binary classification task, i.e. classifying the regions observed by a single-frame data as ground regions or obstacle regions. These approaches might work well in structured environments, however, when extending to complex environments, three problems might occur: (1) The detection results generated by processing a single-frame data is usually unstable and can not provide sufficient coverage for the local environment, especially in complex environments with rough terrains (as shown in Figure \ref{fig_illustration}(b)). (2) The traversability is in itself an ambiguous concept, and is closely related to the vehicle kinematic models. Some special terrains, such as road curbs, ditches, or slopes, might be non-traversable for a small UGV, but can be easily crossed by a larger vehicle. Therefore, instead of trying to classify each region as `ground' or `obstacle', we'd rather accurately model the terrain as well as its normal direction or the travel cost. (3) As the output of the terrain modeling will be the input to the path planning module, we should also distinguish `traversability' from `reachability'. Flat regions surrounded by obstacles, or remote regions that cannot be reached should also be considered as non-traversable regions. In other words, determining traversable regions should not only utilize local information, but also global context.
    
Bearing these principles in mind, we propose a novel LiDAR-based terrain modeling approach for traversability analysis in this paper. It has the following three characteristics: (1) \textbf{Stability}: Terrain modeling is no longer considered as a single-frame task, information from consecutive LiDAR frames is applied for terrain modeling, which could overcome the noise contained in a single frame and promote the stability of results; (2) \textbf{Completeness}: The output of the proposed method is a dense terrain model that provide a dense coverage of the local environment. Properties of those blocked regions are inferred from the neighboring observed regions in real-time; (3) \textbf{Distinguishable travel cost}: Except for those regions that are definitely non-traversable, the rest of the regions are no longer handled by binary classification, but are evaluated with different travel costs according to their geometric characteristics. 

More specifically, we adopt a normal distributions transform (NDT) mapping approach to accurately model the terrain by fusing information from consecutive LiDAR frames. Each grid cell is not only modeled by its mean elevation, but also the variance. To promote the stability and completeness of the generated terrain model, an approach based on spatial-temporal Bayesian generalized kernel (BGK) inference and bilateral filtering is designed to infer the elevation of those unobserved or uncertain regions from the neighboring regions, thus producing a dense and stable terrain model (as shown in Figure \ref{fig_illustration}(c)). Furthermore, terrain traversability analysis is performed by considering the geometric connectivity properties between adjacent terrain regions, and the travel cost is estimated by a local convexity criterion. The travel cost could be utilized to discriminate the traversability of different terrains (as shown in Figure \ref{fig_illustration}(d)). 

We perform experiments on the publicly available SemanticKITTI dataset, RELLIS-3D dataset, and a dataset collected by ourselves. Experimental results show that the proposed method could output accurate, stable and complete terrain models and traversability analysis results in real-time, and outperforms state-of-the-art approaches.  

The rest of this paper is organized as follows. Some related works are introduced in Section \ref{Section: Related_work}, with particular focus on LiDAR-based traversability analysis and terrain modeling approaches. An approach overview and the details of the proposed method are presented in Section \ref{Section: method}. The experimental results and comparisons with state-of-the-art approaches are presented in Section \ref{Section: experiments}. In Section \ref{Section_Conclusion}, the concluding remarks are summarized and our technical contributions are presented.

\section{RELATED WORK}
\label{Section: Related_work}
As surveyed by Papadakis \cite{Papadakis}, LiDAR, camera \cite{Gao}, or proprioceptive sensors \cite{Joonho} are commonly used in traversability analysis task. Among these on-board sensors, proprioceptive sensors only can predict the traversability of regions that have been passed by UGVs; camera is sensitive to illumination changes and cannot work at night. In contrast, LiDAR is an active sensor that can work at night and has a high ranging accuracy, which is ideal for traversability analysis in complex off-road environments. In this paper, we focus on utilizing LiDAR to perform traversability analysis. In general, LiDAR-based traversability analysis algorithms could be roughly classified into two categories: feature-based methods and model-based methods.

\subsection{Feature-based methods}
Feature-based traversability analysis methods mainly utilize the features extracted from point cloud data to determine the terrain traversability. One of the most typical methods is the Min-Max elevation mapping method proposed by Thrun et al. \cite{Thrun}. In this method, the three dimensional (3D) point cloud is firstly projected to form a 2D grid map. The height difference of the points falling in the same grid cell is then utilized to determine the traversability of that grid cell. In addition to the height difference, the average height \cite{Douillard}, the covariance \cite{Hamandi}, the slope \cite{Meng}, or the roughness \cite{Frank} of the grid cell can all be utilized for traversability analysis. These methods are simple to implement and have low computational complexity. However, they usually have limited robustness, and are difficult to adapt to environments with rough terrains.

To improve robustness, some machine learning algorithms could be utilized for traversability analysis. The expectation maximization (EM) algorithm is applied in \cite{Lalonde} to fit a Gaussian mixture model through labeled statistics features, which divides the terrain in forest environment into three categories. A support vector machine (SVM) classifier is trained in \cite{Ahtiainen} to classify the traversable regions from the non-traversable regions, where each grid cell is represented in a NDT traversability map (NDT-TM). A random-forest classifier is trained in \cite{Suger} to detect traversable regions in off-road environment. In \cite{Lee}, a multi-layer perceptron (MLP) model for traversability mapping is trained by self-training algorithm. More recently, Guan et al. \cite{Guan2021TNSTT} fuses some geometric features (the slope and the step height) with semantic features to perform traversability analysis, which is more robust for complex off-road environments.

With the increasing popularity of deep learning, many works have applied it to extract features and perform traversability analysis. To better utilize the convolutional neural network (CNN), researchers come up with two solutions to solve the problem of processing the unordered 3D point cloud. The first solution is to re-encode the point cloud and to produce a dense input tensor for CNN. For example, some works encode the point cloud as an elevation map \cite{Chavez-Garcia}, a 2D birds-eye-view (BEV) map \cite{Shaban}, or a range image \cite{Velas}, and CNNs are trained to perform end-to-end traversable region detection. The other solution is to use PointNet \cite{Charles} based network architecture, which is suitable for unordered 3D point set. An end-to-end approach called GndNet \cite{Paigwar} employs PointNet and Pillar feature encoding network to perform ground segmentation. In addition, some methods try to predict the traversability by the CNNs directly. For instance, Seo et al. \cite{Seo2022ScaTEAS} propose a scalable framework for learning traversability from 3D point cloud in a self-supervised manner. Although these learning-based approaches have achieved good results on some datasets, they usually need a large amount of annotated data for training and the model could not be easily transferred to different environments or LiDAR types, which limits their practical applications. 

In addition, most of existing feature-based traversability analysis methods extract features from a single-frame point cloud. The result is usually unstable across consecutive frames and could not provide a dense coverage of the local environment, especially in complex environments.
 
\subsection{Model-based methods}
For model-based methods, a terrain model of the local environment is first constructed based on the continuity characteristics of the terrain, and then the traversability is determined from the terrain model. One of the most popular approaches is the Bayesian non-parametric inference technique, which assumes the point cloud is spatially correlated, and a continuous terrain surface could be estimated from those observed points. Among the Bayesian non-parametric inference techniques, Gaussian Process Regression (GPR) is widely applied in the traversability analysis task. Directly performing GPR \cite{Vasudevan, Lang} in 2D is a computational expensive task that could hardly meet the real-time requirement of the UGV. To remedy this, Chen et al. \cite{Chen} simplifies the 2D GPR into a series of one-dimensional GPRs. This simplification can reduce the computational complexity. However, the accuracy of the generated terrain model is less accurate. Furthermore, a Variational Hilbert Regression (VHR) based traversability analysis method is proposed in \cite{Guizilini}, where the point cloud is projected into a Reproducing Kernel Hilbert Space (RKHS) and the terrain is approximately estimated in that space. Recently, Bayesian Generalized Kernel (BGK) inference is applied for terrain modeling and traversability analysis. In \cite{Shan}, the traversability analysis is performed online with two sequential BGK inference steps. Although it outperforms VHR and GPR in terms of accuracy and efficiency, the inference procedure only utilizes single-frame point cloud, which is easily influenced by noisy data. Besides, the employed kernel function is too smooth, and can hardly model the terrain discontinuities frequently occurred in complex environments.

Except for Bayesian non-parametric inference, the terrain could also be modeled as a Markov Random Field (MRF) or a Conditional Random Field (CRF). In \cite{Wellington}, the horizontal spatial correlations are encoded in two MRF models, and the traversability inference is performed by Gibbs sampling. In \cite{Zhang}, the terrain smoothness assumption and height measurements are fused in a multi-label MRF model, and loopy belief propagation (BP) algorithm is adopted for inference, providing robust estimation for terrain modeling and ground segmentation. In \cite{Rummelhard}, the terrain is modeled as a spatio-temporal CRF and terrain elevation are estimated by EM algorithm. The graph cut method is used in \cite{Huang} to solve a coarse-to-fine MRF model for ground segmentation. Forkel et al. \cite{Forkel2021} propose a probabilistic terrain estimation approach by modeling it as a recursive Gaussian state estimation problem. This work is further extended as a dynamic resolution version in \cite{Forkel2022}. In addition, B-spline surface \cite{Rodrigues} is also applied in traversability analysis for its powerful surface fitting capability to complex geometrical shapes. 

Some recent works also perform traversability analysis by combining terrain models with vehicle kinematic models. A probabilistic representation of traversability is proposed by Cai et al. \cite{Cai2022}, the traversability is modeled as a distribution conditioned on the dynamical models of robot and the terrain characteristics. Fan et al. \cite{Fan_2021} considers the uncertainty in risk-aware planning and proposes a risk-aware 2.5D traversability evaluation method which accounts for localization error, sensor noise, vehicle model, terrain model, and multiple sources of traversability risk.

\section{METHODOLOGY}
\begin{figure}[t]
	\centering
	\includegraphics[scale=0.35]{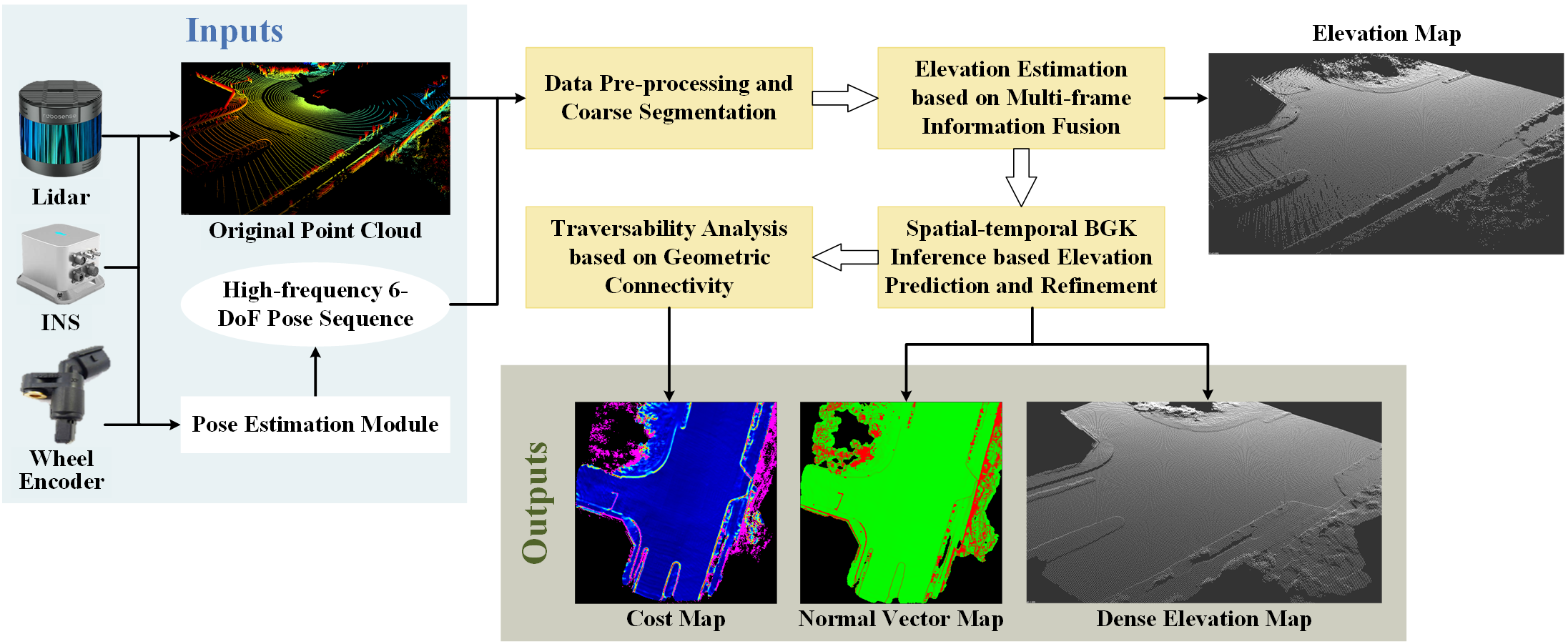} \\
	\caption{The framework of the proposed traversability analysis method. }
	\label{fig_framework}
\end{figure}
\label{Section: method}
The framework of the proposed traversability analysis method is demonstrated in Figure \ref{fig_framework}. The inputs are LiDAR point cloud and high-frequency 6-degree of freedom (DoF) pose sequences obtained from a pose estimation module \cite{Xue2019}, which fuses information from Inertial Navigation System (INS), wheel encoder, and LiDAR odometry. The outputs of the proposed method are a dense terrain model and a cost map of the local surrounding environment. The terrain model is represented as a normal vector map and a dense elevation map.     

The proposed method mainly consists of four modules: a data pre-processing and coarse segmentation module, an elevation estimation module, an elevation prediction and refinement module, and a traversability analysis module. The details of these four modules are described as follows.  

\subsection{Data Pre-processing and Coarse Segmentation}
It is well known that the LiDAR point cloud will be distorted as the UGV moves during the scanning period of one LiDAR scan. To rectify the distorted point cloud, the high frequency 6-DoF poses generated by the pose estimation module are utilized for intra-frame motion compensation. Besides, the point cloud is also rotated to an upright position based on the azimuth, roll and pitch angles obtained from the 6-DoF poses. After this rotation, the terrain properties (such as normal vector, slope, etc) can be well estimated in the axis-aligned coordinate.

The rectified point cloud is then projected into a 2D grid map, and the Min-Max height difference approach \cite{Thrun} is utilized to roughly classify the grid cells into terrain cells and non-terrain cells. In this approach, a height difference threshold $T_h$ needs to be set \textcolor{black}{(the setting of this threshold is listed in Table \ref{tab_parameter})}. A grid cell is considered to be an obstacle cell if its min-max height difference is larger than $T_h$. Besides, the approach introduced in \cite{Jaspers} is also utilized to remove overhanging structures from the point cloud.

\begin{figure}[t]
	\centering
	\includegraphics[scale=0.35]{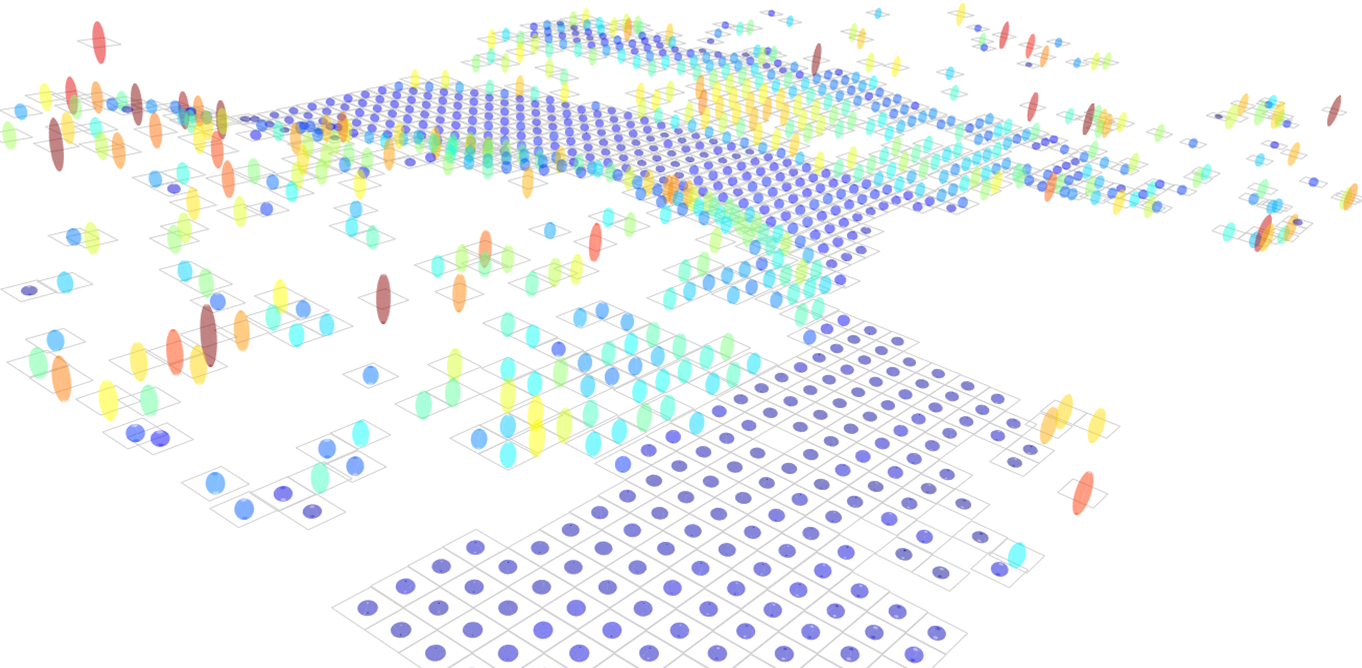} \\
	\caption{An illustrative example of elevation distribution generated from a single LiDAR frame. The elevation of each potential terrain grid cell is represented by a normal distribution (shown as an ellipsoid, a larger radius indicates higher variance).}
	\label{fig_Variance_Single}
\end{figure}

After this coarse classification, the 2D grid map is divided into three parts: the obstacle grid cells, the potential terrain grid cells and the unobserved grid cells. At this stage, it is worth mentioning that the potential terrain grid cells may contain false positives. For example, the roof of nearby vehicles might be mistakenly judged as the potential terrain cells. These mistakes might not be easily corrected by processing a single LiDAR frame, but they can be easily avoided by adopting a multi-frame information fusion strategy to be described in the following subsection. 

For each potential terrain grid cell $\bm{x}_i$, its observed elevation can be modeled as a normal distribution $\mathcal{N}\left(\mu_i, {\Sigma_i}\right)$ by fusing observed heights $\left\{z_{i,j}\right\}_{j = 1:n_i}$ of all $n_i$ points falling in this grid cell:
\begin{equation}
\left\{\begin{split}
\mu_i &= \frac{1}{n_i} \sum_{j=1}^{n_i} z_{i,j} \,, \\
{\Sigma_i} &= \frac{1}{n_i} \sum_{j=1}^{n_i} {z_{i,j}}^2 - {\mu_i}^2 \,,
\label{equation_NDT_observation}
\end{split} \right.
\end{equation}  
where $\mu_i$ and ${\Sigma_i}$ are mean and variance of $\mathcal{N}\left(\mu_i, {\Sigma_i}\right)$, respectively. An illustrative example of elevation distributions generated by a single LiDAR frame is shown in Figure \ref{fig_Variance_Single}.

\subsection{Elevation Estimation based on Multi-frame Information Fusion}
\label{Section: NDT}
\begin{figure}[t]
	\centering
	\includegraphics[scale=0.50]{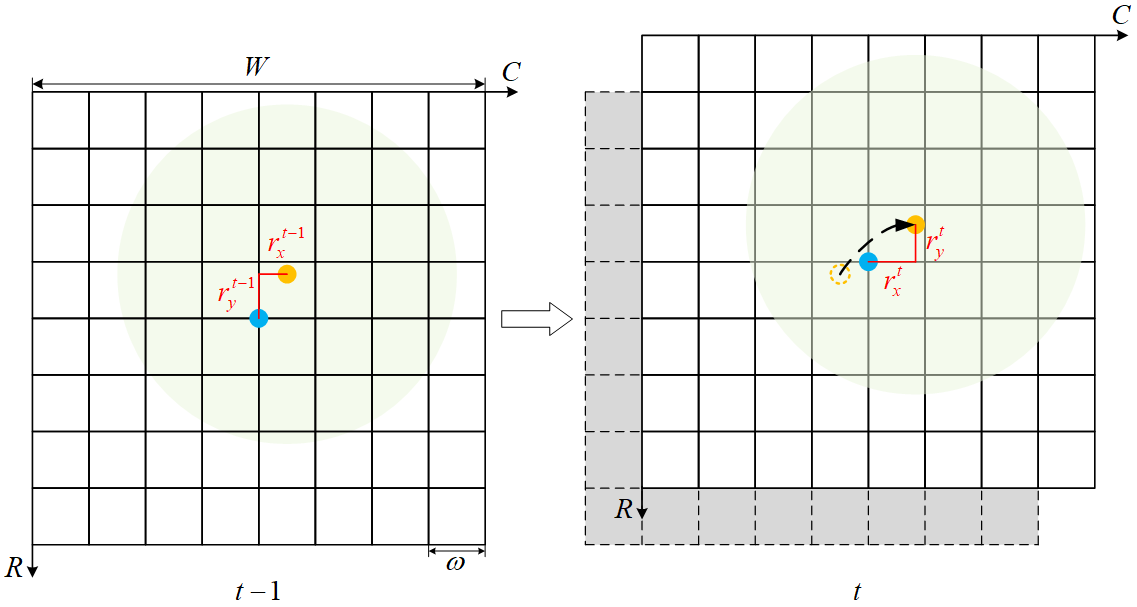} \\
	\caption{The update process of the global grid map. The size of the global grid map is $W \times W$, and the resolution of each grid cell is $\omega \times \omega$. As the UGV moves, the historical grey grid cells exceeding the map boundaries are removed. The shaded green area represents the observation range of the LiDAR. The orange circle denotes the position of the LiDAR, and the blue circle is the center of the grid map.  $\left(r^t_x, r^t_y\right)$ is the calculated residual at time step $t$.}
	\label{fig_NDT}
\end{figure}

To fuse information from consecutive LiDAR frames, the rolling grid technique \cite{Behley} is adopted to construct a global grid map. As illustrated in Figure \ref{fig_NDT}, the size of the global grid map is $W \times W$, and each grid cell represents a $\omega \times \omega$ region, storing the elevation distribution estimated in the previous time step $t-1$ \textcolor{black}{(the settings of $W$ and $\omega$ are listed in Table \ref{tab_parameter})}. As the UGV moves, the historical grid cells exceeding the map boundaries will be removed (shown as the grey cells in Figure \ref{fig_NDT}), and the same number of new cells are generated in the moving direction. 

The center of the global grid map is defined as the bottom left corner of the grid cell where the LiDAR is located in (shown as the blue circle in Figure \ref{fig_NDT}). Let $(L_x^t, L_y^t)$ denote the LiDAR's position represented in the global coordinate system at time step $t$ (shown as the orange circle in Figure \ref{fig_NDT}), then we can compute the residual $\left(r_x^t, r_y^t\right)$ between the LiDAR's global position with respect to the bottom left corner of the grid cell that the LiDAR falls in:
\begin{equation}
\left\{\begin{split}
r_x^t &= L_x^t - \mathrm{Floor}\left(\frac{L_x^t}{\omega}\right) \cdot \omega \,, \\
r_y^t &= L_y^t - \mathrm{Floor}\left(\frac{L_y^t}{\omega}\right) \cdot \omega \,, \\
\end{split}\right.
\label{equation: quantization_error}
\end{equation}
where the function $\mathrm{Floor}\left(a\right)$ is a floor operator that returns the largest integer not greater than $a$. 

As the center of the global map is always defined as the bottom left corner of grid cell that the LiDAR falls in, then the calculated residual exactly represents the translational offset between the LiDAR coordinate and the global map coordinate. Let $(p_x^t, p_y^t, p_z^t)$ denote an observed point represented in the LiDAR coordinate, its global grid coordinate $\left(r, c\right)$ can be calculated as:
\begin{equation}
\left\{ \begin{split}
c &= \frac{W}{2} + \mathrm{Floor}\left(\frac{p^t_x + r_x^t}{\omega}\right) \,, \\
r &= \frac{W}{2} - 1 - \mathrm{Floor}\left(\frac{p^t_y + r_y^t}{\omega}\right) \,.
\end{split}\right.
\label{equation: projection}
\end{equation}

In previous works, the point cloud is usually processed in the LiDAR's local coordinate system where the center of the LiDAR map is aligned with the LiDAR's origin. The output of this process is usually a local grid map that is then fused into the global grid map. In this paper, we emphasize that as the global grid map is discretized by the global coordinate, the LiDAR's local coordinate should be firstly compensated by the calculated residual. With the help of this value, the LiDAR's local grid map will be well aligned with the global grid map, and the quantization error caused by different LiDAR poses can be well eliminated.

Then, information from consecutive frames are fused in the global grid map. For each grid cell $\bm{x}_i$ containing projected points, the observed elevation distribution $\mathcal{N}\left(\mu_i^t, {\Sigma^t_i}\right)$ at timestamp $t$ is firstly calculated by Equation (\ref{equation_NDT_observation}), and the current joint elevation distribution $\hat{\mathcal{N}}\left(\hat{\mu}^t_{i}, {\hat{\Sigma}_{i}^{t}}\right)$ can be estimated by fusing $\mathcal{N}\left(\mu_i^t, {\Sigma^t_i}\right)$ with the previous joint elevation distribution $\hat{\mathcal{N}}\left(\hat{\mu}_i^{t-1}, {\hat{\Sigma}^{t-1}_i}\right)$.

\begin{figure}[t]
	\centering
	\includegraphics[scale=0.42]{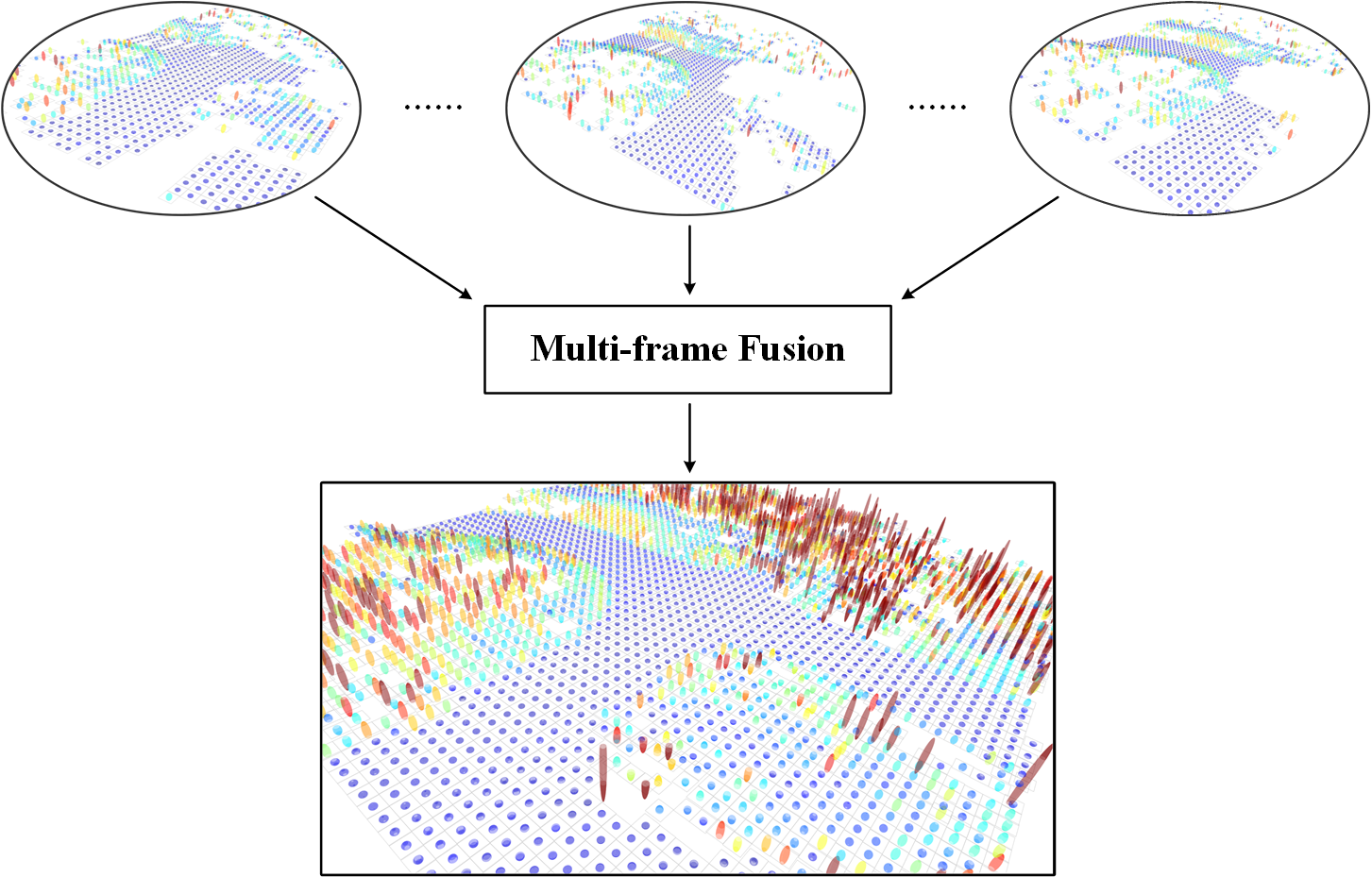} \\
	\caption{An illustrative example showing the elevation distribution estimated by multi-frame information fusion. The elevation distributions generated by each LiDAR frame are fused together in the global grid map.}
	\label{fig_Variance}
\end{figure}

Two fusion strategies could be adopted here: the first one is a Kalman filter based approach that we proposed in our previous work \cite{Xue2021}. In this approach, the observed mean elevation $\mu^t_i$ is used to recursively update the joint elevation distribution $\hat{\mathcal{N}}\left(\hat{\mu}^t_{i}, {\hat{\Sigma}_{i}^{t}}\right)$:
\begin{equation}
\left\{\begin{split}
\bar{\mu}^t_i &= a \hat{\mu}^{t-1}_i \,, \\
\bar{\Sigma}^t_i &= a^2 \hat{\Sigma}^{t-1}_i + \varepsilon \,, \\
K &= \frac{\bar{\Sigma}^t_i c}{c^2\bar{\Sigma}^t_i + \xi} \,, \\
\hat{\mu}^t_i &= \bar{\mu}^t_i + K \cdot \left(\mu_i^t - c \bar{\mu}_i^t\right) \,,\\
\hat{\Sigma}^t_i &= \left(1- Kc\right) \cdot \bar{\Sigma}_i^t \,,
\end{split}\right.
\label{equation:KF}
\end{equation}
where $\bar{\mu}^t_i$ and $\bar{\Sigma}^t_i$ are mean and variance of the prior elevation distribution, $a$ and $c$ are parameters of the prediction model and measurement update model. $K$ is the Kalman gain. $\varepsilon$ and $\xi$ represent the variance of the process noise and measurement noise, respectively. In practice, we found that it is difficult to determine the optimal parameters of $\varepsilon$ and $\xi$ in Equation (\ref{equation:KF}).

In this paper, we propose another fusion strategy based on NDT mapping approach \cite{Saarinen}, which directly fuses the previous joint elevation distribution $\hat{\mathcal{N}}\left(\hat{\mu}_i^{t-1}, {\hat{\Sigma}^{t-1}_i}\right)$ with the observed elevation distribution $\mathcal{N}\left(\mu_i^t, {\Sigma^t_i}\right)$ by:
\begin{equation}
\left\{\begin{split}
S^t_{i} &= S^{t-1}_{i} + n_i^t \,,\\
\hat{\mu}_i^t &= \frac{1}{S^t_{i}} \cdot \left(n_i^t \cdot \mu_i^t + S^{t-1}_{i} \cdot \hat{\mu}^{t-1}_i\right) \,,\\
\hat{\Sigma}_i^t &= \frac{1}{S^t_{i}} \cdot \left[n_i^t \cdot \Sigma_i^t + S^{t-1}_{i} \cdot \hat{\Sigma}^{t-1}_i + \frac{n_i^t \cdot S^{t-1}_{i}}{S^t_{i}} \cdot \left(\mu^t_i - \hat{\mu}^{t-1}_i\right)^2\right]\,,
\label{equation:NDT_joint}
\end{split}\right.
\end{equation}
where $S^t_{i}$ is the total number of projected points in grid cell $\bm{x}_i$ until time step $t$. Compared to the Kalman filter based approach, the above fusion approach makes full use of the mean and variance information contained in each single LiDAR frame, and has no parameters to be set. 

An illustrative example of elevation distribution estimated by multi-frame information fusion is shown in Figure \ref{fig_Variance}, the elevation distribution generated by each single LiDAR frame are fused together in the global grid map, resulting in a more stable and complete result. In addition, the segmentation of potential terrain regions can be refined by stable distribution variance information, if the estimated variance of a grid cell that has been observed multiple times is higher than a variance threshold $T_{\Sigma}$ \textcolor{black}{(the setting of this threshold is listed in Table \ref{tab_parameter})}, this grid cell will be also considered as an obstacle cell.

\subsection{Spatial-temporal BGK Inference based Elevation Prediction and Refinement}
\label{Section: BGK}
The multi-frame information fusion strategy described in previous subsection can improve the stability and completeness of the estimation results. However, there are still some unobserved and unstable regions. To further predict the elevation distribution of those unobserved grid cells and refine the elevation distribution of the other grid cells, an improved spatial-temporal BGK inference approach is proposed. 

The input of the spatial-temporal BGK inference approach is the potential terrain cell set $\bm{O}^t = \left\{\left(\bm{x}_i, \mathcal{N}^t_i\right)\right\}_{i=1:N_o^t}$ obtained in Section \ref{Section: NDT}, where $\mathcal{N}^t_i$ represents the estimated elevation distribution of grid cell $\bm{x}_i$ at time step $t$, and $N_o^t$ is the number of input samples in $\bm{O}^t$. The task is to estimate the elevation distribution $\mathcal{N}_*^t$ of the target grid cell $\bm{x}_*$ based on the input $\bm{O}^t$, which can be further cast as a regression problem and formulated as:
\begin{equation}
p\left(\mathcal{N}_*^t|\bm{x}_*,\bm{O}^t\right) = \int p\left(\mathcal{N}_*^t|\bm{\theta}_*^t\right) \cdot p\left(\bm{\theta}_*^t|\bm{x}_*, \bm{O}^t\right) d\bm{\theta}_*^t\,,
\label{equation:Bayesian_1}
\end{equation}
where $\bm{\theta}_*^t$ is the latent parameter associated with $\bm{x}_*$ at time step $t$.

\textcolor{black}{By applying Bayes' theorem, conditional independence assumption, and a smooth extended likelihood model \cite{Brown} (For details of the derivation, please refer to the ``\emph{Proof 1: The transformation of $p\left(\bm{\theta}_*^t|\bm{x}_*, \bm{O}^t\right)$}" in the Appendix),} $p\left(\bm{\theta}_*^t|\bm{x}_*, \bm{O}^t\right)$ in Equation (\ref{equation:Bayesian_1}) can be further transformed to: 
\begin{equation}
p\left(\bm{\theta}_*^t|\bm{x}_*, \bm{O}^t\right) \propto \prod_{i=1}^{N_o^t} p\left(\mathcal{N}_i^t|\bm{\theta}^t_*\right)^{k\left(\bm{x}_i, \bm{x}_*\right)} \cdot p\left(\bm{\theta}_*^t|\bm{x}_*\right) \,,
\label{equation:Bayesian_4}
\end{equation}
where $k\left(\cdot,\cdot\right)$ is a kernel function. To utilize the estimated variance information $\hat{\Sigma}_i^t$ associated with each grid cell $\bm{x}_i$, we introduce it as a weight in the spatial-temporal BGK inference. Grid cells with larger variance will contribute less in the elevation inference procedure. With the assumptions that the likelihood distribution $p\left(\mathcal{N}_i^t|\bm{\theta}_*^t\right)$ satisfies a normal distribution $\mathcal{N}\left(\tilde{\mu}_i^t, \hat{\Sigma}_i^t\right)$ and the conjugate prior $p\left(\bm{\theta}_*^t|\bm{x}_*\right)$ also satisfies a normal distribution $\mathcal{N}\left(\mu_0, \Sigma_0\right)$, Equation (\ref{equation:Bayesian_4}) can be further represented as:
\begin{equation}
p\left(\bm{\theta}^t_*|\bm{x}_*, \bm{O}^t\right) \propto \prod_{i=1}^{N_o^t} \exp \left[ -\frac{\left(\hat{\mu}_i^t - \tilde{\mu}_i^t\right)^2}{2\hat{\Sigma}_i^t} k\left(\bm{x}_i, \bm{x}_*\right)\right] \exp \left[\frac{-\left(\tilde{\mu}_i^t - \mu_0\right)^2 }{2\Sigma_0}  \right] \,. 
\label{equation:Bayesian_5}
\end{equation}

The mean $\mu_*^t$ and variance ${\Sigma_*^t}$ of $p\left(\bm{\theta}^t_*|\bm{x}_*, \bm{O}^t\right)$ can thus be calculated as:
\begin{equation}
\left\{
\begin{split}
\mu_*^t &= \frac{\sum_{i=1}^{N_o^t} \frac{\hat{\mu}_i^t}{{\hat{\Sigma}_i^t}} k\left(\bm{x}_i, \bm{x}_*\right) + \frac{\mu_0}{\Sigma_0} }{\sum_{i=1}^{N_o^t} \frac{1}{{\hat{\Sigma}_i^t}} k\left(\bm{x}_i, \bm{x}_*\right) + \frac{1}{\Sigma_0}} \,,\\
{\Sigma_*^t} &=  \frac{1}{\sum_{i=1}^{N_o^t} \frac{1}{{\hat{\Sigma}_i^t}} k\left(\bm{x}_i, \bm{x}_*\right) + \frac{1}{\Sigma_0}} \,.
\end{split}
\right.
\label{equation:Bayesian_6}
\end{equation}

\textcolor{black}{For those unobserved grid cells, the mean $\mu_0$ is set to 0 and the variance $\Sigma_0$ is set to positive infinity. For the remaining unstable but observed grid cells, $\mu_0$ and $\Sigma_0$ are equal to the mean $\hat{\mu}_*^t$ and variance $\hat{\Sigma}_*^t$ estimated by Equation (\ref{equation:NDT_joint}).} In addition, the sparse kernel introduced in \cite{Melkumyan} is selected as the kernel function:
\begin{equation}
k\left(\bm{x}_i, \bm{x}_*\right) = \left\{\begin{split}
& \frac{2+\cos\left(2\pi\frac{d_i}{l}\right)}{3}\left(1-\frac{d_i}{l}\right) + \frac{\sin\left(2\pi\frac{d_i}{l}\right)}{2\pi} \quad \left(d_i \leq l\right) \,, \\
& 0 \quad \quad \quad \quad \quad \quad \quad \quad \quad \quad \quad \quad \quad \quad  \quad \quad \quad \,\,\left(d_i>l\right)  \,,
\end{split}\right.
\label{equation:kernel}
\end{equation}
where $l$ is the effective supportive range of the kernel \textcolor{black}{(the setting of this parameter is listed in Table \ref{tab_parameter})}, and $d_i$ denotes the Euclidean distance between $\bm{x}_i$ and $\bm{x}_*$. 

The purpose of the spatial-temporal BGK inference approach is to estimate the elevation $h_*$ of the target grid cell $\bm{x}_*$, and the distribution to be estimated for this regression problem should be $p\left(\mathcal{N}_*^t|\bm{x}_*,\bm{O}^t\right)$ in Equation (\ref{equation:Bayesian_1}). \textcolor{black}{It can be proved that $\mu_*^t$ is also the mean of $p\left(\mathcal{N}_*^t|\bm{x}_*,\bm{O}^t\right)$ (For details of the derivation, please refer to ``\emph{Proof 2: The relationship between $p\left(\bm{\theta}_*^t|\bm{x}_*, \bm{O}^t\right)$ and $p\left(\mathcal{N}_*^t|\bm{x}_*,\bm{O}^t\right)$}" in the Appendix).} Therefore, the elevation $h_*$ of the target grid cell $\bm{x}_*$ can be estimated by Equation (\ref{equation:Bayesian_6}) and (\ref{equation:kernel}).  

In essence, BGK inference could be considered as a Gaussian filtering of the observed potential terrain grid cells. Gaussian filtering is a technique commonly used in image processing area, and a well-known shortcoming of this technique is that edges might be blurred after the filtering. To remedy this, some edge-preserving filtering techniques have been developed. Among these algorithms, bilateral filtering \cite{Tomasi} is a simple and elegant approach that is also suitable for our application scenario. In our case, the edges correspond to regions with sharp terrain changes. 

To perform the bilateral filtering in the spatial-temporal BGK inference, the elevation of each potential terrain cell $\bm{x}_i$ is firstly estimated by Equation (\ref{equation:Bayesian_6}). Then the difference $\delta_i^t$ between the estimated elevation and the observed elevation is calculated. It is reasonable to believe that the difference $\delta_i^t$ will be a large value for grid cells near sharp terrain changes. Therefore, $\delta_i^t$ is transformed to a Gaussian weight $w_i^t$:
\begin{equation}
w_i^t = \exp \left(-\frac{{\delta^t_i}^2}{2\Sigma_w}\right) \,,
\label{equation:bilateral_filtering}
\end{equation}
where $\Sigma_w$ denotes the variance of the Gaussian kernel \textcolor{black}{(the setting of this parameter is listed in Table \ref{tab_parameter})}. This weight is then incorporated into Equation (\ref{equation:Bayesian_6}), and the equation can be further expressed as:
\begin{equation}
\left\{\begin{split}
\mu_*^t &= \frac{\sum_{i=1}^{N_o^t} \frac{\hat{\mu}_i^t}{\hat{\Sigma}_i^t} w_i^t k\left(\bm{x}_i, \bm{x}_*\right) + \frac{\mu_0}{\Sigma_0} }{\sum_{i=1}^{N_o^t} \frac{1}{\hat{\Sigma}_i^t} w_i^t k\left(\bm{x}_i, \bm{x}_*\right) + \frac{1}{\Sigma_0}} \,,\\
\Sigma_*^t &=  \frac{1}{\sum_{i=1}^{N_o^t} \frac{1}{\hat{\Sigma}_i^t}w_i^o k\left(\bm{x}_i, \bm{x}_*\right) + \frac{1}{\Sigma_0}} \,.
\end{split}\right.
\label{equation:Bayesian_7}
\end{equation}

\begin{figure}[t]
	\centering
	\setlength{\abovecaptionskip}{-8pt}
	\subfigure {
		\begin{minipage}[b]{0.4\linewidth}
			\centering
			\includegraphics[scale=0.35]{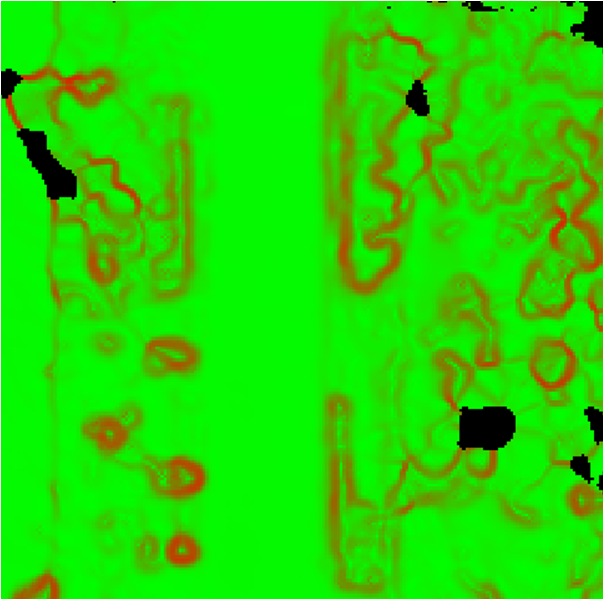}
			\centerline{\footnotesize ({\bf a})}
			\label{fig_B_filter_a}
		\end{minipage}
	} %
	\subfigure{
		\begin{minipage}[b]{0.4\linewidth}
			\centering
			\includegraphics[scale=0.35]{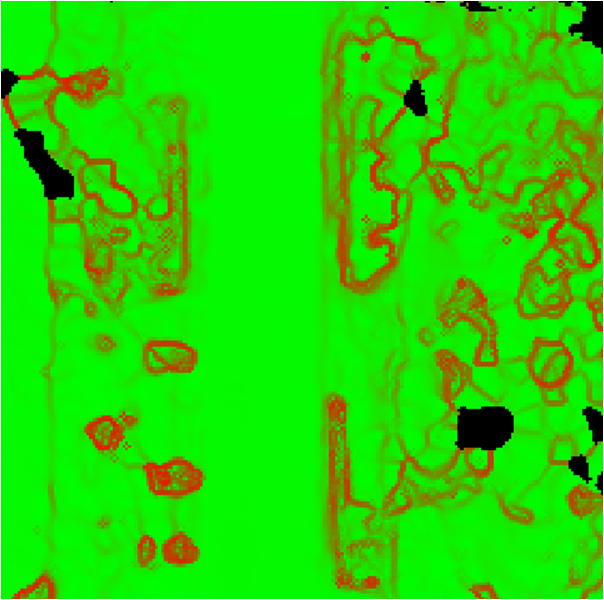}
			\centerline{\footnotesize ({\bf b})}
			\label{fig_B_filter_b}
		\end{minipage}
	} %
	\caption{Normal vector map before (a) and after (b) the bilateral filtering. The color encodes the angle between the estimated normal vector and the vertical axis. A higher degree of greenness indicates a more flat terrain.}
	\label{fig_BF}
\end{figure}

An illustrative example showing the effect of bilateral filtering is demonstrated in Figure \ref{fig_BF}, it is observed that the edges are well retained after bilateral filtering.

\begin{figure}[t]
	\centering
	\includegraphics[scale=0.4]{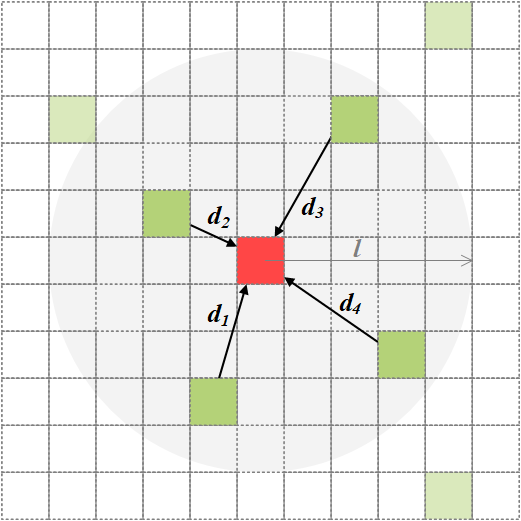} \\
	\caption{An illustration of the BGK inference process. The green cells represent the observed potential terrain grid cells, and the red cell denotes the target grid cell to be estimated. Within the kernel's effective supportive range $l$ (colored by the shaded gray circle), each green grid cell will contribute to the estimation of the red cell, and $d_i$ represents its distance from the red cell.}
	\label{fig_BGK}
\end{figure}

The whole spatial-temporal BGK inference process is illustrated in Figure \ref{fig_BGK}, where the green cells represent the observed potential terrain grid cells, and the red cell denotes the target grid cell to be estimated. Within the kernel's effective supportive range $l$ (colored by the gray circle), each green cell contributes to the estimation of the red cell, and the degree of contribution depends on three factors: its distribution variance $\Sigma_{i}^t$ estimated in Section \ref{Section: NDT}, its distance $d_i$ from the red cell, and the estimation-observation difference $\delta_i^t$ used in the bilateral filtering. 

\subsection{Traversability Analysis based on Geometric Connectivity}
\label{section: Traversability Analysis}
\begin{figure}[h]
	\centering
	\includegraphics[scale=0.45]{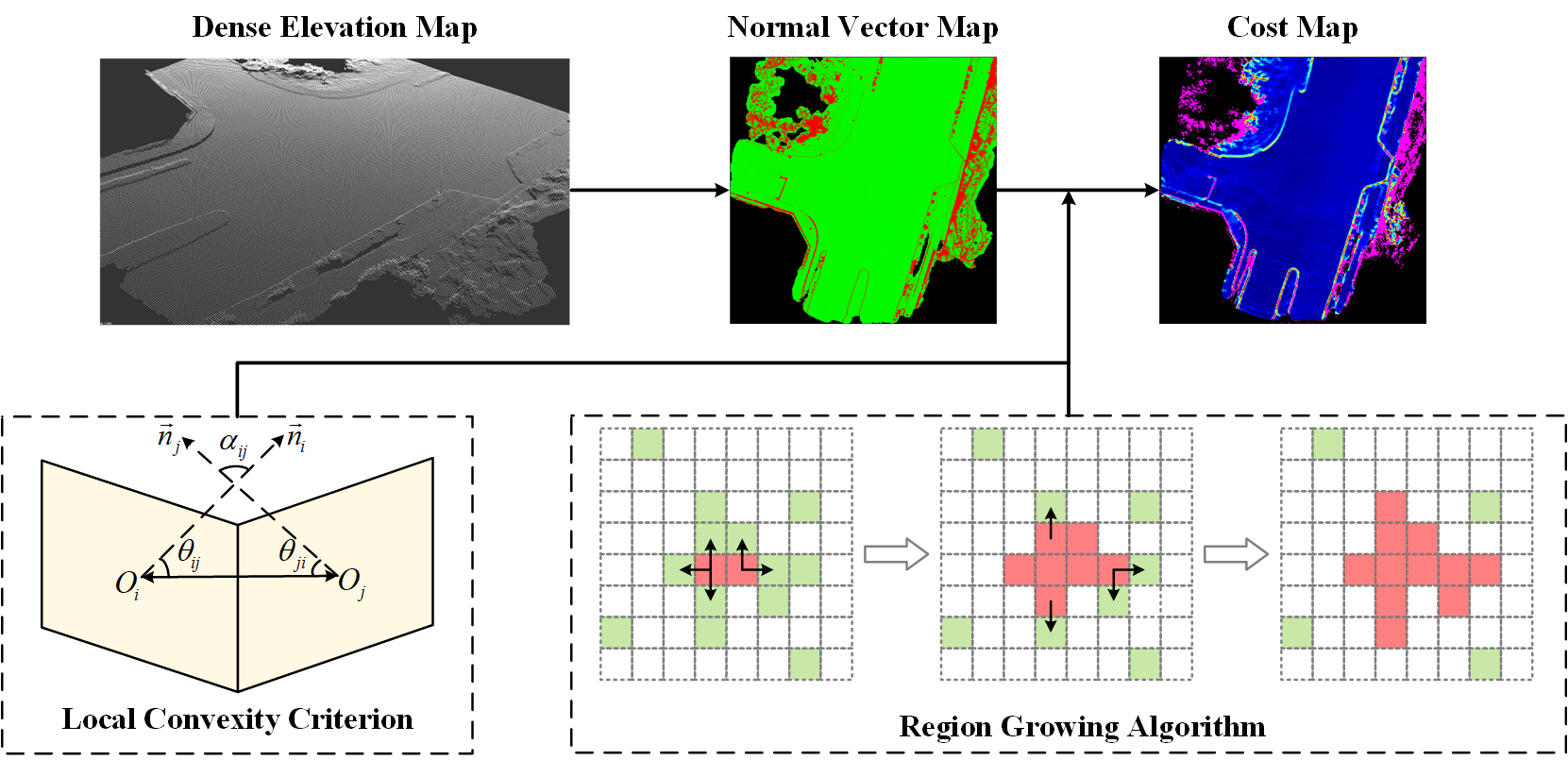} \\
	\caption{The flowchart of terrain traversability analysis. A normal vector map is firstly generated from the dense elevation map. Then, a cost map is calculated by combining the local convexity criterion and the region growing algorithm. For the local convexity criterion, $O_i$ is the center point of the grid cell $\bm{x}_i$, and $\vec{\bm{n}}_i$ is its normal vector, $\alpha_{ij}$ is the angle between the normal vectors $\vec{\bm{n}}_i$ and $\vec{\bm{n}}_j$, $\theta_{ij}$ and $\theta_{ji}$ represent the concavity angles between $\bm{x}_i$ and $\bm{x}_j$.}
	\label{fig_normal}
\end{figure}

After obtaining a dense elevation map from the previous step, terrain traversability can be further analyzed by considering the geometric connectivity properties between adjacent terrain cells. The flowchart of terrain traversability analysis is shown in Figure \ref{fig_normal}. Firstly, a normal vector map is generated from the dense elevation map. Then, the traversability and travel cost of each grid cell are estimated by combining terrain information with local convexity criterion \cite{Moosmann}. Furthermore, starting from the traversable grid cells in the near vicinity of UGV, a region growing operator is performed to determine the terrain reachability, which forms a hard constraint on the final result. The details of these procedures are introduced as follows.

Given a dense elevation map, the normal vector of each grid cell can be calculated by comparing a grid cell's elevation with its neighboring grid cells' elevation. Let $h_{\left(R,C\right)}$ denote the elevation of the grid cell $\left(R,C\right)$ in the grid map, and its 3D coordinate can be expressed as $\vec{\bm{v}}_{\left(R,C\right)} = \left(x_R,y_C,h_{\left(R,C\right)}\right)$. Then, the normal vector $\vec{\bm{n}}_{\left(R,C\right)}$ of the grid cell $\left(R,C\right)$ can be calculated by:
\begin{equation}
\vec{\bm{n}}_{\left(R,C\right)} = \frac{\left(\vec{\bm{v}}_{\left(R,C+1\right)} - \vec{\bm{v}}_{\left(R,C-1\right)}\right) \times \left(\vec{\bm{v}}_{\left(R-1,C\right)} - \vec{\bm{v}}_{\left(R+1,C\right)}\right)}{\left\Vert \left(\vec{\bm{v}}_{\left(R,C+1\right)} - \vec{\bm{v}}_{\left(R,C-1\right)}\right) \times \left(\vec{\bm{v}}_{\left(R-1,C\right)} - \vec{\bm{v}}_{\left(R+1,C\right)}\right)\right \Vert}\,, 
\label{equation:normal_vector}
\end{equation}
where $\vec{\bm{v}}_{\left(R,C+1\right)}$, $\vec{\bm{v}}_{\left(R,C-1\right)}$,  $\vec{\bm{v}}_{\left(R-1,C\right)}$, and $\vec{\bm{v}}_{\left(R+1,C\right)}$ are 3D coordinates of 4-neighborhoods of grid cell $\left(R,C\right)$, `$\times$' represents the cross product operator.

With the estimated dense elevation map and normal vector map, the traversability of each grid cell can be analyzed by local convexity criterion \cite{Moosmann}. As shown in Figure \ref{fig_normal}, if an UGV can pass through two adjacent gird cells $\bm{x}_i$ and $\bm{x}_j$, the angle $\alpha_{ij}$ between their corresponding normal vectors $\vec{\bm{n}}_i$ and $\vec{\bm{n}}_j$ should be small, and the level of concavity (represented by concavity angles $\theta_{ij}$ and $\theta_{ji}$) between two grid cells should not be too high. Given the maximum similarity angle $T_{\alpha}$ and the minimum concavity angle $T_{\theta}$ according to the kinematic constraints of UGV \textcolor{black}{(the settings of $T_{\alpha}$ and $T_{\theta}$ are listed in Table \ref{tab_parameter})}, the path between gird cells $\bm{x}_i$ and $\bm{x}_j$ is considered to be traversable if it satisfies:
\begin{equation}
\left\{\frac{\vec{\bm{n}}_i \cdot \vec{\bm{v}}_{ij}}{\left\Vert \vec{\bm{v}}_{ij}\right\Vert} \leq \cos\left(T_{\theta}\right)\right\}\, \cap  \,  \left\{\frac{\vec{\bm{n}}_j \cdot \vec{\bm{v}}_{ji}}{\left\Vert \vec{\bm{v}}_{ji}\right\Vert} \leq \cos\left(T_{\theta}\right) \right\} \, \cap  \, \left\{ \vec{\bm{n}}_i \cdot \vec{\bm{n}}_j \geq \cos\left(T_{\alpha}\right) \right\}\,,
\label{equation:geometric connectivity}
\end{equation}
where $\vec{\bm{v}}_{ij} = \vec{\bm{v}}_j - \vec{\bm{v}}_i$ is the  displacement vector from $\vec{\bm{v}}_i$ to $\vec{\bm{v}}_j$, and $\vec{\bm{v}}_{ji} = -\vec{\bm{v}}_{ij}$. The travel cost $C_i$ of each traversable grid cell $\bm{x}_i$ can be computed as:
\begin{equation}
C_i = \frac{1}{3m} \sum_{j=1}^{m} \left[\frac{\vec{\bm{n}}_i \cdot \vec{\bm{v}}_{ij}}{\left\Vert \vec{\bm{v}}_{ij}\right\Vert \cos\left(T_{\theta}\right)} + \frac{\vec{\bm{n}}_j \cdot \vec{\bm{v}}_{ji}}{\left\Vert \vec{\bm{v}}_{ji}\right\Vert \cos\left(T_{\theta}\right)} + \frac{\cos\left(T_{\alpha}\right)}{\vec{\bm{n}}_i \cdot \vec{\bm{n}}_j}\right] \,,
\label{equation:travel_cost}
\end{equation}
where $m$ is the total number of traversable neighboring cells ($m\leq4$).

Finally, taking the terrain reachability into account, a region growing algorithm is performed. As shown in Figure \ref{fig_normal}, the seed cells of the traversable regions are firstly selected in the near vicinity of UGV by considering the LiDAR's installation height. Starting from these seed cells, any of its 4-neighboring grid cells satisfying Equation (\ref{equation:geometric connectivity}) are appended to the seed set. The algorithm terminates if the seed set stops growing. All the grid cells in the seed set are considered as the final traversable regions.

In this way, except for those definitely non-traversable and unreachable regions, the remaining terrain regions are all associated with a travel cost value stored in a cost map. This cost map can discriminate the traversability of different terrains and provide richer information to the path planning module, which is helpful for UGV to autonomously select reasonable and safe driving paths in complex outdoor environments.

\section{EXPERIMENTAL RESULTS AND DISCUSSIONS}
\label{Section: experiments}

\subsection{Data Preparation}
Experiments are conducted on three different datasets: the SemanticKITTI dataset \cite{SemanticKITTI} collected in urban environment, the RELLIS-3D dataset \cite{Rellis3D} collected in rugged off-road environment, and our own dataset collected in both urban and off-road scenarios.

\subsubsection{Dataset Description}
\label{section_Dataset}
SemanticKITTI dataset contains point-wise semantic annotations from 28 semantic classes for all 22 LiDAR sequences of the KITTI Odometry Benchmark \cite{KITTI}. The data were collected from inner city, residential areas, highway scenes or countryside roads around Karlsruhe, Germany. This dataset also provides ground-truth poses for each point cloud frame. In our experiments, we select sequences 00, 05 and 07 for testing (denoted as S-00, S-05, S-07, respectively). Detailed information about these sequences are presented in Table \ref{tabel_data}.

RELLIS-3D dataset is a multimodal dataset contains annotations for 13,556 LiDAR scans. All sequences were recorded in the Ground Research facility on the Rellis Campus of Texas A\&M University, which is a rugged off-road environment that includes different kinds of terrains and obstacles. All LiDAR scans are point-wise labeled with 20 different classes, and the ground-truth poses for each point cloud frame are also provided by a high-precision Simultaneous Localization And Mapping (SLAM) system. For our experiments, sequences 00, 01 and 02 are selected for testing (denoted as R-00, R-01, R-02, respectively), and the detailed information about these sequences are presented in Table \ref{tabel_data}.
	
\begin{figure}[t]
	\centering
	\includegraphics[scale=0.5]{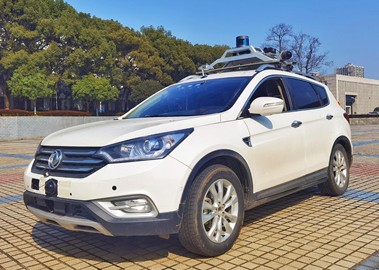} \\
	\caption{Our experimental vehicle. It is equipped with a RS-Ruby128 LiDAR, a StarNeto XW-GI7660 GNSS/INS system, and two wheel encoders.}
	\label{fig_UGV}
\end{figure}

\begin{figure}[H]
	\centering
	\includegraphics[scale=0.28]{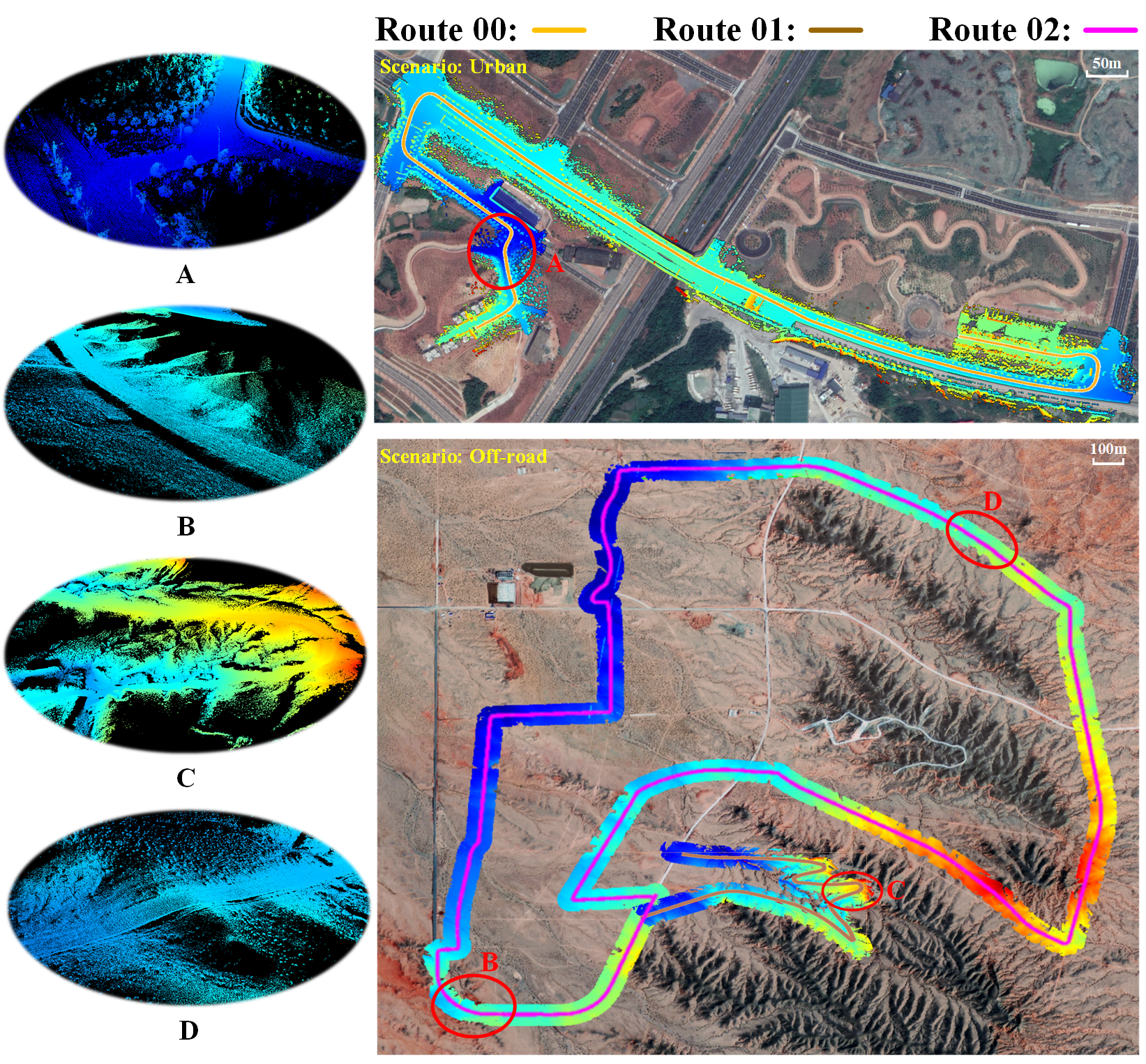} \\
	\caption{Our own dataset consists of three testing routes. Their corresponding pre-build High Definition (HD) maps are overlaid on the satellite images. The HD-maps are colored by elevation. Four typical scenes (locations A, B, C,and D) in the HD-maps are shown on the left.} 
	\label{fig_Mapping}
\end{figure}

Our own dataset was collected by our own experimental vehicle in both urban and off-road environments. As shown in Figure \ref{fig_UGV}, a Robosense RS-Ruby128 LiDAR is mounted on the top of the vehicle, which has 128 laser beams with a maximum detection distance of 250 meters. The horizontal Field of View (FoV) is $360^{\circ}$, and the vertical FoV is $40^{\circ}$. Besides the LiDAR, the vehicle is also equipped with a StarNeto XW-GI7660 GNSS/INS system and two wheel encoders. Based on our previous work \cite{Xue2019}, we can obtain high frequency 6-DoF poses at centimeter-level accuracy. As shown in Figure \ref{fig_Mapping}, three testing routes are selected for testing (denoted as O-00, O-01, O-02, respectively). Route O-00 was collected in the urban environment with paved roads, and the other two routes were collected in the desert and mountain environments with more challenging uneven dirt tracks. Route O-01 has the most uneven terrain, and Route O-02 has the longest trajectory length and the largest elevation gain. Detailed information about these testing routes are presented in Table \ref{tabel_data}.

\begin{table}[H]
	\centering
	\caption{The detailed information of testing routes}  
	\begin{tabular}{cccccc} 
		\toprule
		\textbf{Dataset} &
		\tabincell{c}{\textbf{Route} \\ \textbf{ID}} & \tabincell{c}{\textbf{Trajectory} \\ \textbf{length} (km)}& \tabincell{c}{\textbf{Duration}\\ \textbf{time} (min)} & \tabincell{c}{\textbf{Average}\\ \textbf{speed} (km/h)} & \tabincell{c}{\textbf{Elevation}\\ \textbf{gain} (m)} \\
		\midrule
		\multirow{3}{*}{SemanticKITTI} & S-00 & 3.71 & 7.84 & 28.40 & 22.18 \\
		& S-05 & 2.20 & 4.79 & 27.58 & 22.74 \\
		& S-07 & 0.69 & 1.91 & 21.84 & 4.30 \\
		\multirow{3}{*}{Rellis-3D} & R-00 & 0.33 & 4.74 & 4.20 & 6.90 \\
		& R-01 & 0.27 & 3.86 & 4.16 & 6.04 \\
		& R-02 & 0.48 & 6.91 & 4.19 & 14.85 \\ 	
		\multirow{3}{*}{Ours} & O-00 & 1.67 & 3.69 & 27.24 & 15.30  \\
		& O-01 & 2.18 & 5.82 & 22.41 & 55.51 \\
		& O-02 & 10.99 & 26.56 & 24.83 & 230.58 \\
		\bottomrule	
	\end{tabular}
	\label{tabel_data}
\end{table} 

\subsubsection{Ground-truth Generation}
To evaluate the performance of the proposed method, the ground-truth of the traversability map and the terrain elevation map are required. The whole process to generate the ground-truth consists of the following four steps.

\begin{figure}[t]
	\centering 
	\setlength{\abovecaptionskip}{-8pt}
	\subfigure {
		\begin{minipage}[b]{25ex}
			\centering
			\includegraphics[scale=0.27]{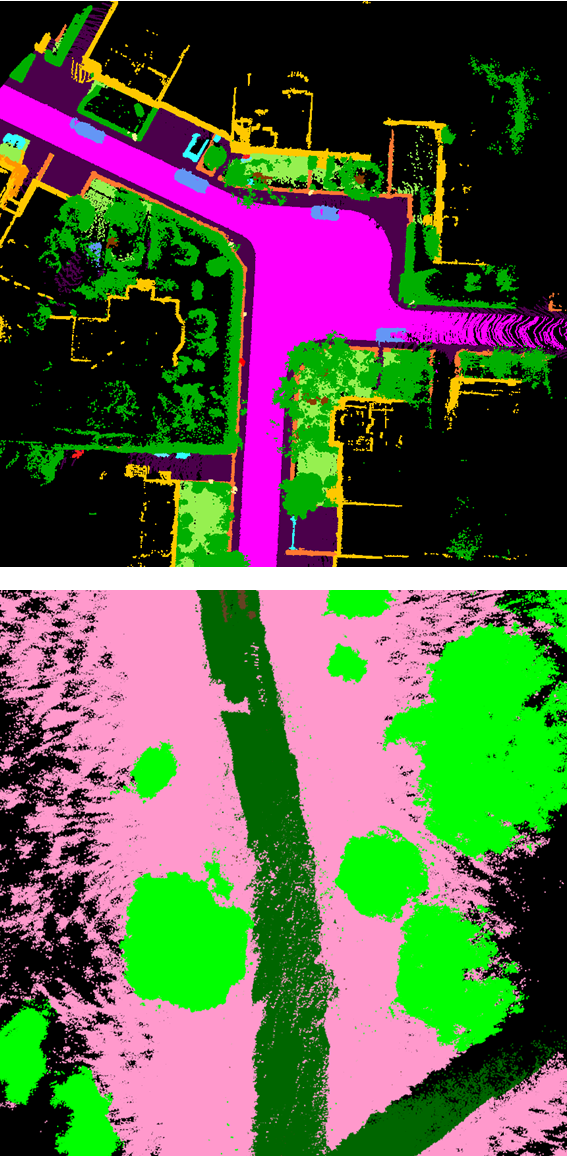}
			\centerline{\footnotesize ({\bf a})}
			\label{fig_GT_a}
		\end{minipage}
	} %
	\subfigure{
		\begin{minipage}[b]{30ex}
			\centering
			\includegraphics[scale=0.27]{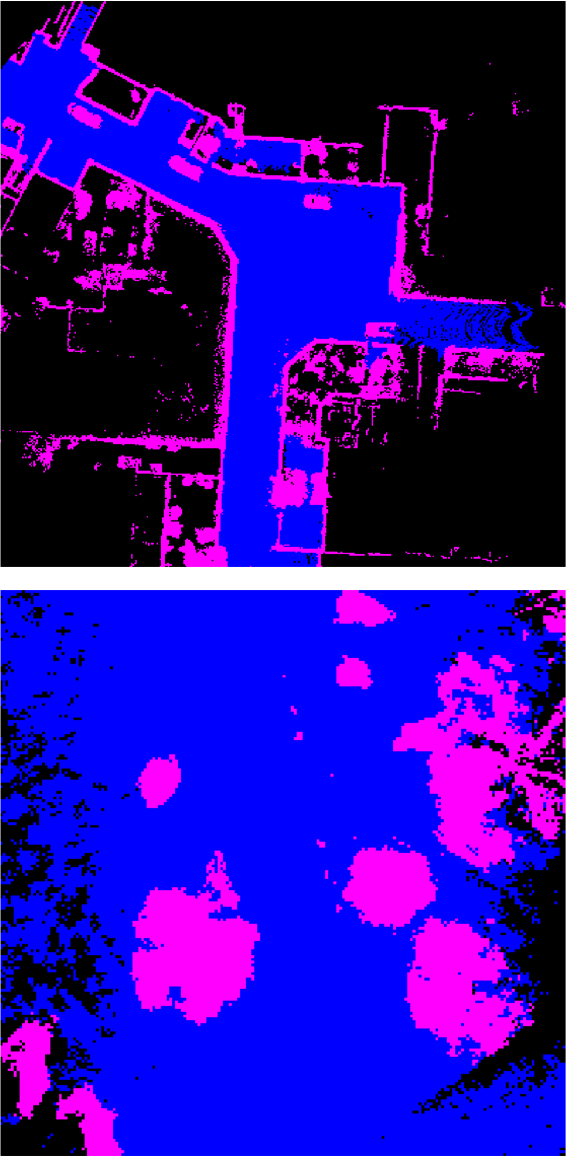}
			\centerline{\footnotesize ({\bf b})}
			\label{fig_GT_b}
		\end{minipage}
	} %
	\subfigure{
		\begin{minipage}[b]{40ex}
			\centering
			\includegraphics[scale=0.27]{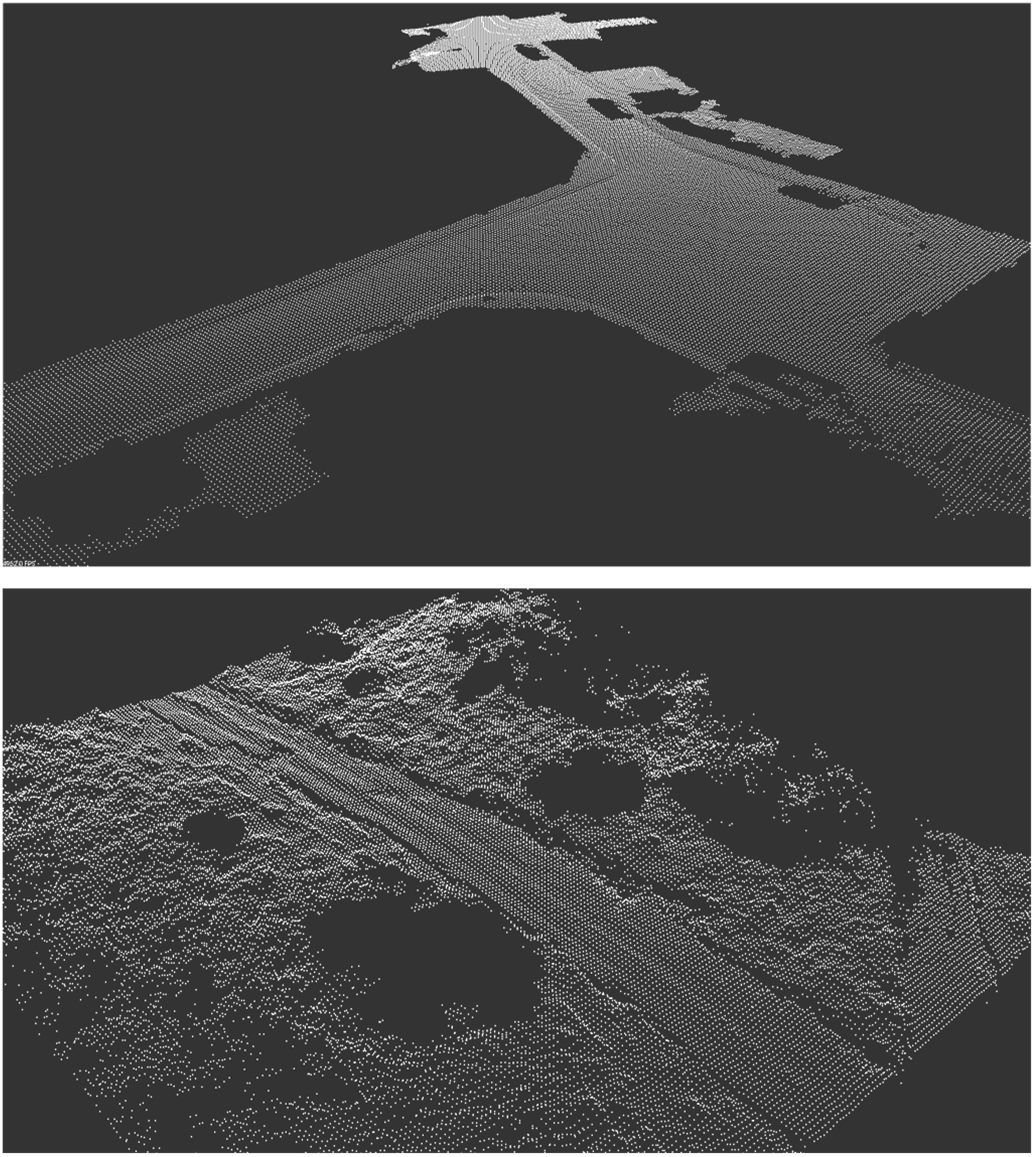}
			\centerline{\footnotesize ({\bf c})}
			\label{fig_GT_c}
		\end{minipage}
	} %
	\caption{The ground-truth of traversability and terrain elevation generated for SemanticKITTI (the first row) and RELLIS-3D (the second row). (a) is the assembled semantic map, where different colors represent different semantic labels; (b) shows the ground-truth of traversability (pink regions denote non-traversable cells and blue regions represent traversable cells); (c) shows the ground-truth of terrain elevation. }
	\label{fig_GT}
\end{figure}

\textbf{Step 1: Assembling local frames to obtain a dense map}. To generate the ground-truth for point cloud $\bm{f}_{i}$, we firstly assemble local frames around $\bm{f}_{i}$ to generate a dense map. A frame $\bm{f}_{k_i}$ is selected if its distance to $\bm{f}_{i}$ is smaller than a distance threshold. Each point $\bm{p}_{k_i}$ in $\bm{f}_{k_i}$ is then transformed to the local coordinate of $\bm{f}_{i}$ by:
\begin{equation}
\tilde{\bm{p}}_{k_i} = {\bm{T}_{i}}^{-1} \cdot \bm{T}_{k_i} \cdot \bm{p}_{k_i} \,,
\label{equation_label}
\end{equation}
where $\tilde{\bm{p}}_{k_i}$ denotes the coordinate of $\bm{p}_{k_i}$ after the transformation. $\bm{T}_{i}$ and $\bm{T}_{k_i}$ are the transformation matrices calculated from the frame poses of $\bm{f}_{i}$ and $\bm{f}_{k_i}$, respectively.

\textbf{Step 2: Traversability judgement}. The merged point cloud is projected into a 2D grid map (as shown in Figure \ref{fig_GT_a}), and the traversability $L^{G}_{j}$ of each grid cell $\bm{x}_j$ is determined by the semantic labels associated with points $\bm{P}_{j}$ that falling in the grid cell $\bm{x}_j$. If all of the semantic labels of $\bm{P}_{j}$ are within the set of traversable labels, the grid cell $\bm{x}_j$ is considered as a traversable cell (shown as blue regions in Figure \ref{fig_GT_b}); otherwise, the grid cell $\bm{x}_j$ is considered as a non-traversable cell (shown as pink regions in Figure \ref{fig_GT_b}).

\textbf{Step 3: Region growing}. The region growing method described in Section \ref{section: Traversability Analysis} is performed on the generated traversability map, and the initial seed cell is set as the center of traversability map. The purpose of this operation is to filter out those traversable regions that are unreachable to the UGV. After this step, the ground-truth of the traversability map can be generated.

\textbf{Step 4: Terrain elevation calculation}. The terrain elevation $H_{j}^{\mathrm{G}}$ of each traversable grid cell $\bm{x}_j$ can be calculated by averaging the z-values of all the projected points $\bm{P}_{j}$ in this grid cell (as shown in Figure \ref{fig_GT_c}). After this step, the ground-truth of the terrain elevation map can be generated.

In SemanticKITTI, the traversable labels include `road', `parking', `sidewalk', `other-ground', and `terrain', whilst in RELLIS-3D, the traversable labels include `grass', `asphalt', `log', `bush', `concrete', `puddle', `mud' and `rubble'. For those points labeled as `vegetation' in SemanticKITTI dataset or `tree' in RELLIS-3D dataset, we determine if they are hanging obstacles based on their distances from those ground points projected in the same grid cell. A `tree' point or a `vegetation' point is considered to be hanging obstacle if the distance between its height and the maximum height of all ground points in the same grid cell is larger than a threshold $T_{o}$ ($T_o$ depends on the height $h_u$ of the UGV, and $T_o$ is taken as $h_u+0.5$ (m) in this paper). The points judged as hanging obstacles are not involved in the traversability judgement.

For our own dataset, since there is no semantic label, we only perform step 1 and 4 to generate the ground-truth of the terrain elevation map. The frame poses of our dataset are proved by the offline mapping approach proposed in \cite{Ren}, and the accurate 3D high definition (HD) maps of three routes are shown in Figure \ref{fig_Mapping}.

\subsection{Evaluation Metrics}
The proposed method is evaluated from two perspectives: namely the traversability estimation and the terrain elevation estimation. Three performance indicators (precision $P$, recall $R$ and F-measure $F_1$) are used to evaluate the traversability estimation results. $P$ reflects the traversability estimation accuracy, and can be expressed as the true positive ratio of all traversable grid cells in the estimated result; $R$ reflects the completeness performance, and can be expressed as the coverage rate of traversable grid cells from the ground-truth; $F_1$ is a combined performance indicator. These three performance indicators are defined as: 
\begin{equation}
\left\{\begin{split}
P &= \frac{1}{N_{E}} \sum\limits_{\textcolor{black}{j=1}}^{\textcolor{black}{N_{E}}}\left(L^E_{j} \cap L^{G}_{j} \right) \,, \\
R &= \frac{1}{N_{G}} \sum\limits_{\textcolor{black}{j=1}}^{\textcolor{black}{N_{G}}}\left(L^E_{j} \cap L^G_{j} \right) \,, \quad  L^E_{j},L^G_{j} \in \left\{0,1\right\} \\
F_1 &= \frac{2 P R}{P+R} \,,
\end{split}\right.
\label{equation:indices}
\end{equation} 
where $N_{E}$ and $N_{G}$ denote the total number of traversable grid cells in an estimated traversable grid map and its corresponding ground-truth, respectively. \textcolor{black}{All $N_{E}$ traversable grid cells in the estimated traversability map are counted for calculating $P$, and all $N_{G}$ traversable grid cells in the ground-truth map are counted for calculating $R$.}  $L^E_{j}$ and $L^{G}_{j}$ represent the traversability label of the $j$-th grid cell in the estimation result and its corresponding ground-truth, and $L^E_{j},L^G_{j} \in \left\{0,1\right\}$.

To evaluate the terrain elevation estimation results, we use the terrain coverage rate $R_c$ and the root mean squared error $E$ as performance metrics. Whilst $R_c$ represents the completeness of the recovered terrain model, $E$ reflects the model accuracy. These two performance indicators are defined as:
\begin{equation}
\left\{\begin{split}
R_c &= \frac{1}{N_{G}} \sum\limits_{\textcolor{black}{j=1}}^{\textcolor{black}{N_{G}}}\left[V\left(H^E_{j}\right) \cap V\left(H^{G}_{j} \right)\right]  \,, \\
E &= \frac{1}{N_{G}}\sum_{\textcolor{black}{j=1}}^{\textcolor{black}{N_{G}}} \sqrt{\left(H_{j}^E - H_{j}^{G}\right)^2} \,,
\end{split} \right.
\label{equation:RMSE}
\end{equation}
\textcolor{black}{where all $N_{G}$ traversable grid cells in the ground-truth map are counted for calculating $R_c$ and $E$.} $H_{j}^E$ is the estimated terrain elevation of the $j$-th grid cell and $H_{j}^{G}$ is its corresponding ground-truth, $V(H)$ is a Boolean function that judges if the elevation $H$ of a grid cell is valid or not. This function returns 0 if $H$ is equal to an initial default value (this value is set as $-999$ in this paper, representing the grid cell contains no valid elevation value); otherwise, it returns 1.

The terrain elevation estimation is evaluated on all the three datasets, whilst the traversability estimation is only evaluated on SemanticKITTI and RELLIS 3D dataset. The parameter settings of the proposed method are presented in Table \ref{tab_parameter}, and they remain unchanged during the experiments. All experiments are performed on a laptop with a 2.20GHz Intel Core i7-8750H CPU and 32 GB main memory, no GPU parallel computation or multi-threading tricks are utilized for boosting speed.
\begin{table}[H]
	\centering
	\caption{Parameter settings of the proposed method.}  
	\begin{tabular}{ccccccccc}
		\toprule
		\textbf{Parameter} & $W$ & $\omega$ & $T_h$ & $T_{\Sigma}$ &\textcolor{black}{ $\Sigma_w$} & $l$ & $T_{\alpha}$ & $T_{\theta}$ \\
		\midrule
		\textbf{Value} &  80$\,$m & 0.2$\,$m & 0.4$\,$m & 0.1 & \textcolor{black}{0.1} & 1$\,$m & $10^{\circ}$ & $80^{\circ}$  \\
		\bottomrule
	\end{tabular}
	\label{tab_parameter}
\end{table}

\begin{figure}[t]
	\centering 
	\subfigure {
		\begin{minipage}[b]{\linewidth}
			\centering
			\includegraphics[scale=0.385]{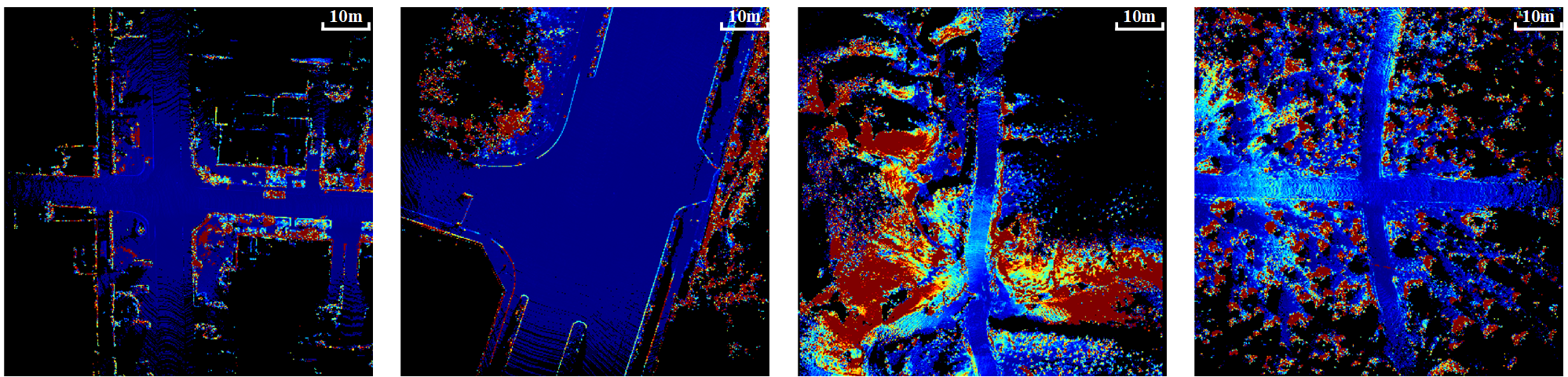}
			\centerline{\footnotesize ({\bf a}) NDT mapping}
			\vspace{-0.2in} 
			\label{fig_Variance_a}
		\end{minipage}
	} \vspace{-0.2in} 
	\subfigure{
		\begin{minipage}[b]{\linewidth}
			\centering
			\includegraphics[scale=0.385]{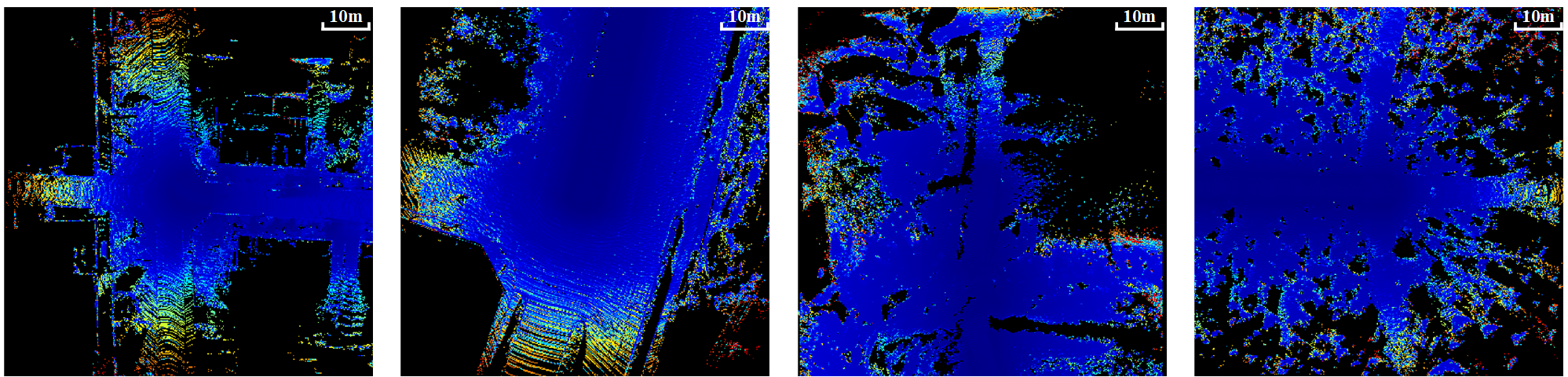}
			\centerline{\footnotesize ({\bf b}) KF}
			\label{fig_Variance_b}
		\end{minipage}
	} 
	\subfigure{
		\begin{minipage}[b]{\linewidth}
			\centering
			\includegraphics[scale=0.385]{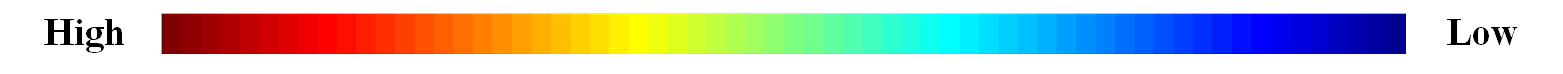}
			\label{fig_Variance_color}
		\end{minipage}
	} %
	\caption{The variance maps generated by different multi-frame information fusion strategies. (a) and (b) are results generated by NDT mapping approach and Kalman filter method respectively. The estimated variance for each grid cell is visualized by different colors.}
	\label{fig_variacne}
\end{figure}

\begin{table}[t]
	\centering
	\caption{Performance comparison under different multi-frame information fusion strategies.}  
	\begin{tabular}{cccccccc} 
		\toprule
		\tabincell{c}{\textbf{Route} \\ \textbf{ID}} & \tabincell{c}{\textbf{Fusion} \\ \textbf{Strategy}} & \tabincell{c}{$\bm{P}$ \\ (\%)} &  \tabincell{c}{$\bm{R}$ \\ (\%)} & \tabincell{c}{$\bm{F_1}$  \\ (\%)} & \tabincell{c}{$\bm{E}$\\ (cm)} & \tabincell{c}{$\bm{R_c}$ \\ (\%)} & \tabincell{c}{\textbf{Time} \\(ms)} \\
		\midrule
		\multirow{2}{*}{S-00} & KF & \textBF{98.42} & 70.74 & 82.31 & 2.84  & 81.83 & 43.01 \\
		& NDT & 97.72 & \textBF{75.79} & \textBF{85.37} & \textBF{2.37} & 81.83 & \textBF{40.10} \\
		\cline{1-8} 
		\multirow{2}{*}{S-05} & KF & \textBF{98.41} & 70.75 & 82.31 & 2.71 & 80.22  & 43.70 \\
		& NDT & 97.08 & \textBF{75.93} & \textBF{85.21} & \textBF{2.27} & 80.22 & \textBF{41.30} \\
		\cline{1-8} 
		\multirow{2}{*}{S-07} & KF & \textBF{98.65} & 71.24 & 82.74 & 3.04 & 83.06 & 45.96 \\
		& NDT & 97.66 & \textBF{77.15} & \textBF{86.20} & \textBF{2.63} & 83.06 & \textBF{43.11} \\
		\cline{1-8} 
		\multirow{2}{*}{R-00} & KF & \textBF{97.94} & 57.13 & 72.16 & 5.36 & 94.19 & 49.16 \\
		& NDT & 97.38 & \textBF{62.28} & \textBF{75.97} & \textBF{4.46} & 94.19 & \textBF{48.83} \\
		\cline{1-8} 
		\multirow{2}{*}{R-01} & KF & \textBF{99.76} & 71.36 & 83.20 & 7.81 & 92.70 & 48.02 \\
		& NDT & 99.69 & \textBF{73.61} & \textBF{84.69} & \textBF{7.28} & 92.70 & \textBF{47.80} \\
		\cline{1-8} 
		\multirow{2}{*}{R-02} & KF & \textBF{99.94} & 71.38 & 83.28 & 4.94 & 94.50 & 53.65 \\
		& NDT & 99.89 & \textBF{72.59} & \textBF{84.08} & \textBF{4.27} & 94.50 & \textBF{52.38} \\
		\cline{1-8} 
		\multirow{2}{*}{O-00} & KF & \multicolumn{3}{c}{\multirow{2}*{\diagbox[width=8em]{\,}{\,}}} & 1.91 & 96.56 & 69.02  \\
		& NDT &  &  &  & \textBF{1.45} & 96.56 & \textBF{65.41} \\
		\cline{1-8} 
		\multirow{2}{*}{O-01} & KF & \multicolumn{3}{c}{\multirow{2}*{\diagbox[width=8em]{\,}{\,}}} &  4.74 & 91.97 &  70.19 \\
		& NDT &  &  &  & \textBF{4.27} & 91.97 & \textBF{66.77} \\
		\cline{1-8}  
		\multirow{2}{*}{O-02} & KF &  \multicolumn{3}{c}{\multirow{2}*{\diagbox[width=8em]{\,}{\,}}} &   3.88 & 94.66  & 79.80     \\
		& NDT &  &  &  & \textBF{3.56} & 94.66 & \textBF{74.40}  \\
		\bottomrule	
	\end{tabular}
	\label{tabel_Fusion_Strategies}
\end{table} 
\subsection{Evaluation on Multi-frame Information Fusion Strategies}
\label{Section: Fusion_Strategy}
We firstly perform experiments to compare the performance of two different multi-frame information fusion strategies presented in Section \ref{Section: NDT}. The parameters used in the Kalman Filter method are set as follows: $a$ and $c$ in Equation (\ref{equation:KF}) are set to 1, the variance $\varepsilon$ of the process noise is set to $0.01$, and the variance $\xi$ of the measurement noise is set to $0.01d_i$, where $d_i$ is the distance between the grid cell and the LiDAR origin. In contrast, there are no other parameters used in the NDT-mapping approach. Quantitative experimental results are shown in Table \ref{tabel_Fusion_Strategies}, and the variance maps generated by these two fusion strategies are compared in Figure \ref{fig_variacne}, where the estimated variance for each grid cell is visualized by different colors.

$\mathbf{Discussion}$: From Figure \ref{fig_Variance_a}, it can be observed that regions with sharp terrain changes usually have a larger variance value. This variance information is utilized in the NDT mapping approach, but ignored in the Kalman filter method. Therefore, the results obtained by the Kalman filter method contains much more noise than the NDT mapping approach, and this can be reflected in Table \ref{tabel_Fusion_Strategies}. It shows that the NDT mapping approach obtains a higher score in most performance indicators ($R$, $F_1$, and $E$) than the Kalman filter method on all three datasets. In summary, this experiment demonstrates that the NDT mapping approach is a more preferred choice for multi-frame information fusion than the Kalman filter method.

\subsection{Evaluation on Spatial-temporal BGK Elevation Inference}
\label{Section: BGK_Inference}
For the Spatial-temporal BGK elevation inference, we make two improvements over the baseline method \cite{Shan} in Section \ref{Section: BGK}. The first one is to introduce bilateral filtering to BGK elevation inference, thus alleviating the edge blurring problem caused by the Gaussian filtering. The second improvement is that the estimated variance of each grid cell is introduced as a weight in the BGK elevation inference. With the help of this weight, grid cells with larger variance will contribute less to BGK elevation inference.

\begin{figure}[H]
	\centering 
	\setlength{\abovecaptionskip}{-8pt}
	\subfigure {
		\begin{minipage}[b]{0.23\linewidth}
			\centering
			\includegraphics[scale=0.38]{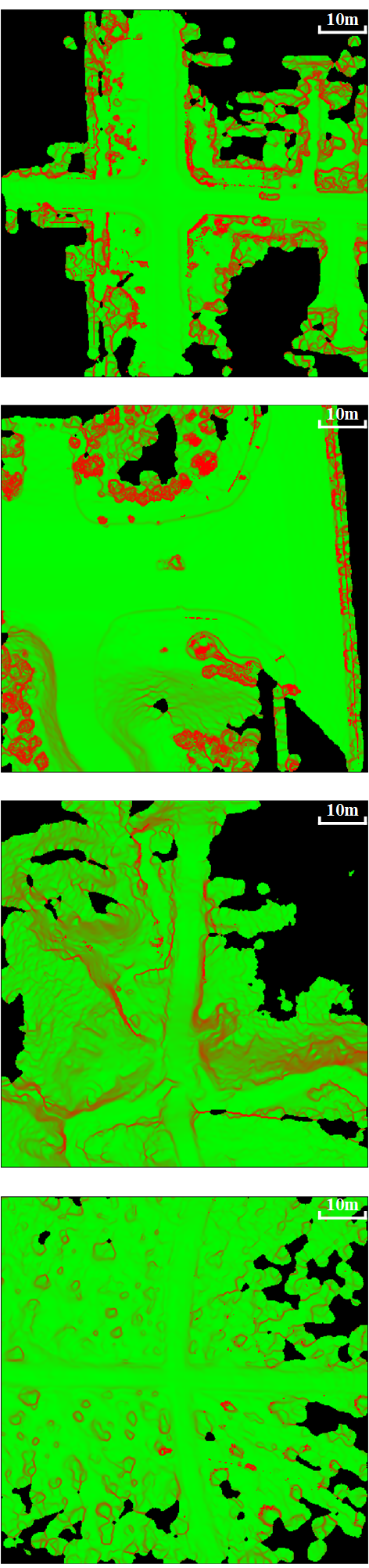}
			\centerline{\footnotesize ({\bf a}) EV $+$ BF}
			\label{fig_Contrast_BGK_a}
		\end{minipage}
	} %
	\subfigure{
		\begin{minipage}[b]{0.23\linewidth}
			\centering
			\includegraphics[scale=0.38]{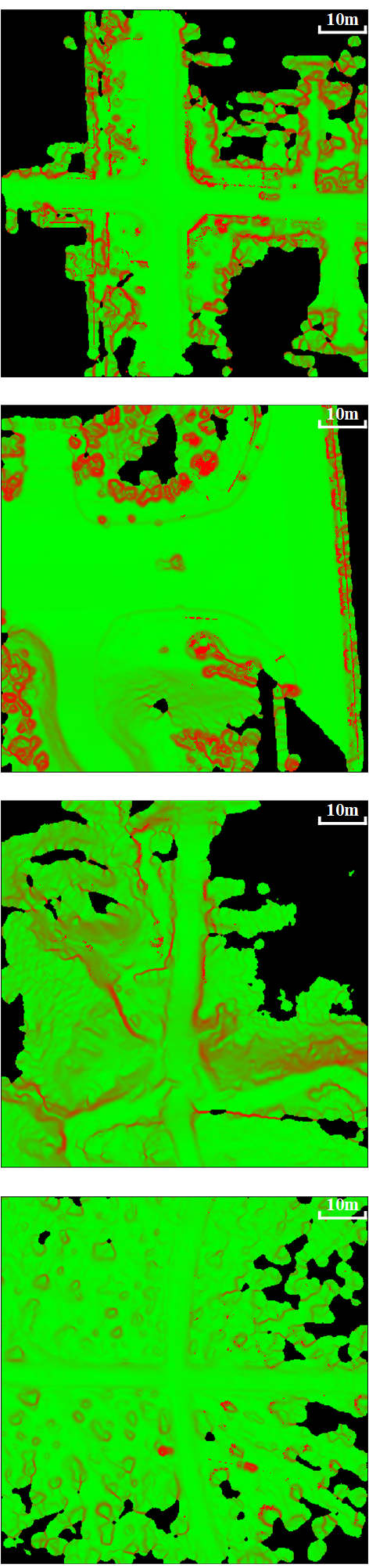}
			\centerline{\footnotesize ({\bf b}) EV $+$ $\neg$BF}
			\label{fig_Contrast_BGK_b}
		\end{minipage}
	} %
	\subfigure{
		\begin{minipage}[b]{0.23\linewidth}
			\centering
			\includegraphics[scale=0.38]{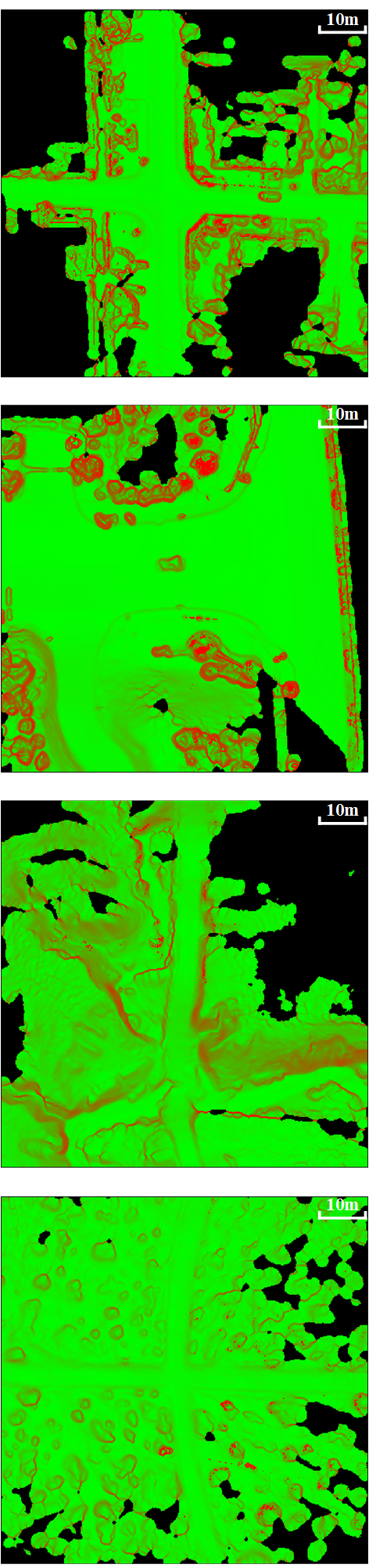}
			\centerline{\footnotesize ({\bf c}) $\neg$EV $+$ BF}
			\label{fig_Contrast_BGK_c}
		\end{minipage}
	} %
	\subfigure{
		\begin{minipage}[b]{0.23\linewidth}
			\centering
			\includegraphics[scale=0.38]{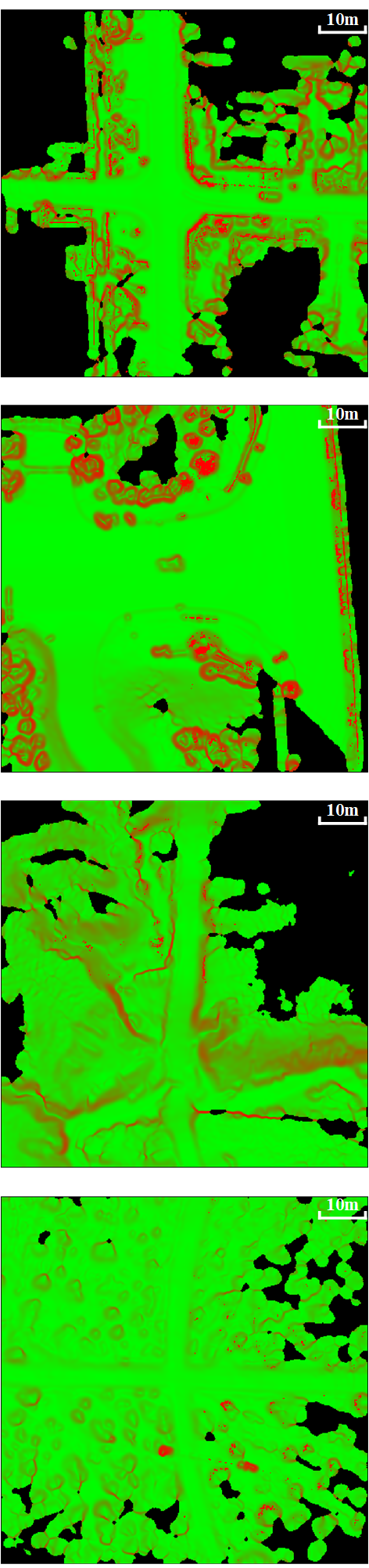}
			\centerline{\footnotesize ({\bf d}) $\neg$EV $+$ $\neg$BF}
			\label{fig_Contrast_BGK_d}
		\end{minipage}
	} %
	\caption{The normal vector maps obtained under different settings of the BGK elevation inference: ( (a) Both the estimated variance (EV) and the bilateral filtering (BF) modules are introduced; (b) Only the EV is introduced, the BF is not utilized; (c) Only the BF is introduced, the EV is set to a constant value; (d) The EV is set to a constant value, and the BF is not utilized). The color of each grid cell encodes the angle between the estimated normal vector and the vertical axis, a higher degree of greenness indicates a more flat terrain.} 
	\label{fig_Compare_BGK}
\end{figure}

To verify the effectiveness of the added bilateral filtering module, comparative experiments are conducted with and without the bilateral filtering (BF) module. Then, comparative experiments are also performed when adding the estimated variance (EV) as a weight in the BGK elevation inference, or simply set the weight  to a constant value. The
estimated variance refers to $\hat{\Sigma}_i^t$ in Equation (\ref{equation:Bayesian_7}). In all these experiments, the NDT mapping method is applied as the multi-frame information fusion approach. Quantitative experimental results are shown in Table \ref{tabel_BGK}. The normal vector maps obtained under different settings of BGK elevation inference are compared in Figure \ref{fig_Compare_BGK}, where the color of each grid cell encodes the angle between the estimated normal vector and the vertical axis, a higher degree of greenness indicates a more flat terrain.

$\mathbf{Discussion}$: From Figure \ref{fig_Compare_BGK}, it can be observed that the results shown in Figure \ref{fig_Contrast_BGK_b} and \ref{fig_Contrast_BGK_d} are more blurred than those in Figure \ref{fig_Contrast_BGK_a} and \ref{fig_Contrast_BGK_c}, which confirms the effectiveness of the added bilateral filtering module. In addition, in Figure \ref{fig_Contrast_BGK_a} and \ref{fig_Contrast_BGK_b}, the regions with sharp terrain changes (the red regions) are smaller compared to those in Figure \ref{fig_Contrast_BGK_c} and \ref{fig_Contrast_BGK_d}. This can prove that the introduced estimated variance in the BGK elevation inference can reduce the estimated errors caused by those regions with sharp terrain changes. In addition, quantitative experimental results in Table \ref{tabel_BGK} suggests that both the performance of traversability estimation and terrain elevation estimation are improved by applying these two improvements in the BGK elevation inference. In conclusion, this experiment demonstrates that the two improvements proposed in the Spatial-temporal BGK elevation inference are critical for accurate terrain modeling and traversability analysis.

\subsection{Evaluation on Grid Cell Resolution}
\label{Section: Fineness}
To evaluate how different grid resolutions might influence the algorithm performance, the comparative experiments are performed in this subsection. Three different grid cell resolutions ($0.1 \mathrm{m} \times 0.1 \mathrm{m}$, $0.2 \mathrm{m} \times 0.2 \mathrm{m}$, and $0.4 \mathrm{m} \times 0.4 \mathrm{m}$) are selected for performance comparison. Quantitative experimental results are presented in Table \ref{tabel_Fineness}.

$\mathbf{Discussion}$: The results in Table \ref{tabel_Fineness} indicate that the coarsest grid cell resolution ($0.4 \mathrm{m} \times 0.4 \mathrm{m}$) has the fastest computation time, but the worst performance. For the finest grid resolution $0.1 \mathrm{m} \times 0.1 \mathrm{m}$, the performance metrics $P$ and $E$ are the highest, but the performance metrics $R$, $R_c$ and $F_1$ are lower than the results obtained by the medium grid cell resolution $0.2 \mathrm{m} \times 0.2 \mathrm{m}$, and its computational complexity is much higher. Compromising between accuracy and computation time, we select $0.2 \mathrm{m} \times 0.2 \mathrm{m}$ as the most suitable grid cell resolution in the proposed method.

\begin{table}[H]
	\centering
	\caption{Performance comparison under different settings of the BGK elevation inference.}  
	\begin{tabular}{ccccccccc} 
		\toprule
		\tabincell{c}{\textbf{Route} \\ \textbf{ID}} & \tabincell{c}{\textbf{Estimated} \\ \textbf{Variance}} & \tabincell{c}{\textbf{Bilateral} \\ \textbf{Filter}} & \tabincell{c}{$\bm{P}$ \\ (\%)} &  \tabincell{c}{$\bm{R}$ \\ (\%)} & \tabincell{c}{$\bm{F_1}$  \\ (\%)} & \tabincell{c}{$\bm{E}$\\ (cm)} & \tabincell{c}{$\bm{R_c}$ \\ (\%)} & \tabincell{c}{\textbf{Time} \\(ms)} \\
		\midrule
		\multirow{4}{*}{S-00} & \multirow{2}{*}{$\times$} & $\times$ & 93.73 & 70.04 & 80.17 & 4.07 &  81.83 & \textBF{30.54} \\
		&  & $\surd$ & 95.69 & 71.71 & 81.98 & 2.82 & 81.83 & 37.84  \\
		& \multirow{2}{*}{$\surd$} & $\times$ & 96.46 & 73.85 & 83.65 & 2.61 & 81.83 & 31.55 \\
		&  & $\surd$ & \textBF{97.72} & \textBF{75.79} & \textBF{85.37} & \textBF{2.37} & 81.83 & 40.10 \\
		\cline{1-9} 
		\multirow{4}{*}{S-05} & \multirow{2}{*}{$\times$} & $\times$ & 93.29 & 68.76 & 79.17 & 4.12 & 80.22 & \textBF{31.28} \\
		&  & $\surd$ & 95.15 & 71.21 & 81.46 & 2.75 & 80.22 & 39.53  \\
		& \multirow{2}{*}{$\surd$} & $\times$ & 95.62 & 73.36 & 83.02 & 2.53 & 80.22 & 32.58 \\
		&  & $\surd$ & \textBF{97.08} & \textBF{75.93} & \textBF{85.21} & \textBF{2.27} & 80.22 & 41.30 \\
		\cline{1-9} 
		\multirow{4}{*}{S-07} & \multirow{2}{*}{$\times$} & $\times$ & 93.38 & 71.91 & 81.25 &  4.70 & 83.06 & \textBF{31.39}\\
		&  & $\surd$ & 95.54 & 73.60 & 83.15 & 3.17 & 83.06 & 38.26 \\
		& \multirow{2}{*}{$\surd$} &  $\times$ & 96.31 & 75.96 & 84.93 & 2.88 & 83.06 & 32.52  \\
		&  & $\surd$ & \textBF{97.66} & \textBF{77.15} & \textBF{86.20} & \textBF{2.63} & 83.06 & 43.11 \\
		\cline{1-9} 
		\multirow{4}{*}{R-00} & \multirow{2}{*}{$\times$} & $\times$ & 91.94 & 58.69 & 71.65 & 9.61 & 94.19 & \textBF{36.10}  \\
		&  & $\surd$ & 94.61 & 60.70 & 73.95 & 6.56 & 94.19 & 45.14 \\
		& \multirow{2}{*}{$\surd$} & $\times$ & 95.63 & 61.56 & 74.90 & 5.86 & 94.19 & 38.62  \\
		&  & $\surd$ & \textBF{97.38} & \textBF{62.28} & \textBF{75.97} & \textBF{4.46} & 94.19 & 48.83  \\
		\cline{1-9} 
		\multirow{4}{*}{R-01} & \multirow{2}{*}{$\times$} & $\times$ & 99.09 & 71.38 & 82.98 & 9.36 & 92.70 &  \textBF{36.59} \\
		&  & $\surd$ & 99.37 & 71.94 & 83.46 & 7.78 & 92.70 & 45.79  \\
		& \multirow{2}{*}{$\surd$} & $\times$ & 99.40 & 72.16 & 83.62 & 8.34 & 92.70 & 38.57 \\
		&  & $\surd$ & \textBF{99.69} & \textBF{73.61} & \textBF{84.69} & \textBF{7.28} & 92.70 & 47.80   \\
		\cline{1-9} 
		\multirow{4}{*}{R-02} & \multirow{2}{*}{$\times$} & $\times$ & 99.82 & 70.08 & 82.35 & 6.17 & 94.50 & \textBF{43.14} \\
		&  & $\surd$ & 99.86 & 71.03 & 83.01 & 5.25 & 94.50 & 49.45 \\
		& \multirow{2}{*}{$\surd$} & $\times$ & 99.86 & 71.56 & 83.37 & 5.78 & 94.50 & 43.20 \\
		&  & $\surd$ & \textBF{99.89} & \textBF{72.59} & \textBF{84.08} & \textBF{4.27} & 94.50 & 52.38
		\\
		\cline{1-9} 
		\multirow{4}{*}{O-00} & \multirow{2}{*}{$\times$} & $\times$ &  \multicolumn{3}{c}{\multirow{4}*{\diagbox[width=8em]{\,}{\,}}} & 3.15 & 96.56 & \textBF{45.97} \\
		&  & $\surd$ &  &  &  & 1.83 & 96.56 & 62.97 \\
		& \multirow{2}{*}{$\surd$} & $\times$ &  &  &  &  1.76 & 96.56 & 47.42 \\
		&  & $\surd$ &  &  &  & \textBF{1.45} & 96.56 & 65.41 \\
		\cline{1-9} 
		\multirow{4}{*}{O-01} & \multirow{2}{*}{$\times$} & $\times$ &  \multicolumn{3}{c}{\multirow{4}*{\diagbox[width=8em]{\,}{\,}}} & 5.80 & 91.97 & \textBF{47.69} \\
		&  & $\surd$ &  &  &  & 4.71 & 91.97 & 63.38  \\
		& \multirow{2}{*}{$\surd$} & $\times$ &  &  &   & 4.58 & 91.97 & 49.30 \\
		&  & $\surd$ &  &  &  & \textBF{4.27} & 91.97 & 66.77 \\
		\cline{1-9}  
		\multirow{4}{*}{O-02} & \multirow{2}{*}{$\times$} & $\times$ &  \multicolumn{3}{c}{\multirow{4}*{\diagbox[width=8em]{\,}{\,}}} & 5.37 & 94.66 & \textBF{52.42} \\
		&  & $\surd$ &  &  &  & 3.92 & 94.66 & 71.65 \\ 
		& \multirow{2}{*}{$\surd$} & $\times$ &  &  &   & 3.76 & 94.66 & 53.32  \\
		&  & $\surd$ &  &  &  & \textBF{3.56} & 94.66 & 74.40 \\
		\bottomrule	
	\end{tabular}
	\vspace{-1.8em}
	\label{tabel_BGK}
\end{table} 
\begin{table}[t]
	\centering
	\caption{Performance comparison under different grid cell resolutions.}  
	\begin{tabular}{cccccccc} 
		\toprule
		\tabincell{c}{\textbf{Route} \textbf{ID}} & \tabincell{c}{\textbf{Grid} \textbf{Resolution}} & \tabincell{c}{$\bm{P}$ (\%)} &  \tabincell{c}{$\bm{R}$ (\%)} & \tabincell{c}{$\bm{F_1}$  (\%)} & \tabincell{c}{$\bm{E}$ (cm)} & \tabincell{c}{$\bm{R_c}$ (\%)} & \tabincell{c}{\textbf{Time} (ms)} \\
		\midrule
		\multirow{3}{*}{S-00} & 0.1m $\times$ 0.1m & \textBF{97.96} & 73.08 & 83.71 & \textBF{1.89} & 77.79 & 167.39 \\
		& 0.2m $\times$ 0.2m & 97.72 & \textBF{75.79} & \textBF{85.37} & 2.37 & \textBF{81.83} & 40.10 \\
		& 0.4m $\times$ 0.4m & 97.55 &  69.40 & 81.10 & 3.26 & 79.75 & \textBF{10.81} \\
		\cline{1-8} 
		\multirow{3}{*}{S-05} & 0.1m $\times$ 0.1m & \textBF{98.06} & 74.50 & 84.67 & \textBF{1.96} & 76.95 & 182.46  \\
		& 0.2m $\times$ 0.2m & 97.08 & \textBF{75.93} & \textBF{85.21} & 2.27 & \textBF{80.22} & 41.30 \\
		& 0.4m $\times$ 0.4m & 97.29 & 68.51 & 80.40  &  3.06 & 77.54 & \textBF{11.28} \\
		\cline{1-8} 
		\multirow{3}{*}{S-07} & 0.1m $\times$ 0.1m & \textBF{97.89} & 74.67 & 84.71 & \textBF{2.14} & 77.24 & 182.37  \\
		& 0.2m $\times$ 0.2m & 97.66 & \textBF{77.15} & \textBF{86.20} & 2.63 & \textBF{83.06} & 43.11
		\\
		& 0.4m $\times$ 0.4m & 97.50 & 70.26 & 81.67 & 3.43 & 80.33 & \textBF{10.87} \\
		\cline{1-8} 
		\multirow{3}{*}{R-00} & 0.1m $\times$ 0.1m & \textBF{97.52} & 61.05 & 75.09 & \textBF{4.17} & 92.27 & 197.72  \\
		& 0.2m $\times$ 0.2m & 97.38 & \textBF{62.28} & \textBF{75.97} & 4.46 & \textBF{94.19} & 48.83  \\
		& 0.4m $\times$ 0.4m & 97.21 & 59.83 & 74.07 & 4.82 & 93.35 & \textBF{15.68} \\
		\cline{1-8} 
		\multirow{3}{*}{R-01} & 0.1m $\times$ 0.1m & \textBF{99.82} & 72.90 & 84.26 & \textBF{7.06} & 91.86 & 193.95 \\
		& 0.2m $\times$ 0.2m & 99.69 & \textBF{73.61} & \textBF{84.69} & 7.28 & \textBF{92.70} & 47.80  \\
		& 0.4m $\times$ 0.4m & 99.51 & 72.07 & 83.60 & 7.61 & 92.17 & \textBF{13.99} \\
		\cline{1-8} 
		\multirow{3}{*}{R-02} & 0.1m $\times$ 0.1m & \textBF{99.93} & 72.14 & 83.79 & \textBF{4.11} & 93.18 & 215.73  \\
		& 0.2m $\times$ 0.2m & 99.89 & \textBF{72.59} & \textBF{84.08} & 4.27 & \textBF{94.50} & 52.38  \\
		& 0.4m $\times$ 0.4m & 99.85 & 71.32 & 83.21 & 4.93 & 93.96 & \textBF{19.82} \\
		\cline{1-8} 
		\multirow{3}{*}{O-00} & 0.1m $\times$ 0.1m &    \multicolumn{3}{c}{\multirow{3}*{\diagbox[width=8em]{\,}{\,}}} & \textBF{1.31} & 94.28 & 373.55  \\
		& 0.2m $\times$ 0.2m &  &  &  & 1.45 & \textBF{96.56} & 65.41 \\
		& 0.4m $\times$ 0.4m &  &  &  & 1.99 & 95.04 & \textBF{15.33} \\
		\cline{1-8} 
		\multirow{3}{*}{O-01} & 0.1m $\times$ 0.1m & \multicolumn{3}{c}{\multirow{3}*{\diagbox[width=8em]{\,}{\,}}} & \textBF{3.86} & 88.98 & 381.86 \\
		& 0.2m $\times$ 0.2m &  &  &  & 4.27 & \textBF{91.97} & 66.77 \\
		& 0.4m $\times$ 0.4m &  & & & 4.99 & 89.66 & \textBF{15.29} \\
		\cline{1-8}  
		\multirow{3}{*}{O-02} & 0.1m $\times$ 0.1m &  \multicolumn{3}{c}{\multirow{3}*{\diagbox[width=8em]{\,}{\,}}} & \textBF{3.32} & 92.08 & 392.05  \\
		& 0.2m $\times$ 0.2m &  &  &  & 3.56 & \textBF{94.66} & 74.40  \\
		& 0.4m $\times$ 0.4m & & & & 4.21 & 93.39 & \textBF{15.83} \\
		\bottomrule	
	\end{tabular}
	\label{tabel_Fineness}
\end{table}   
\subsection{Comparison with State-of-the-art Methods}
\label{Section_comparison}
In this subsection, the proposed method is compared with three typical non-learning-based traversability analysis methods (a Gaussian Process Regression (GPR) based approach \cite{Chen}, a NDT terrain mapping (NDT-TM) based approach \cite{Ahtiainen}, and a BGK inference based approach \cite{Shan}) and a learning-based traversability analysis method (GndNet \cite{Paigwar}). 

For GndNet, we only compare it on SemanticKITTI dataset since it was only trained on this dataset. Besides, this method could only produce a terrain model at a coarse resolution of $1 \mathrm{m} \times 1 \mathrm{m}$, we have to upsample its result to $0.2 \mathrm{m} \times 0.2 \mathrm{m}$ for quantitative comparison. For the other methods, the experiments are performed on all three datasets and the grid cell resolution is set to $0.2 \mathrm{m} \times 0.2 \mathrm{m}$.
\begin{figure}[H]
	\centering 
	\subfigure {
		\begin{minipage}[b]{0.48\linewidth}
			\centering
			\includegraphics[scale=0.28]{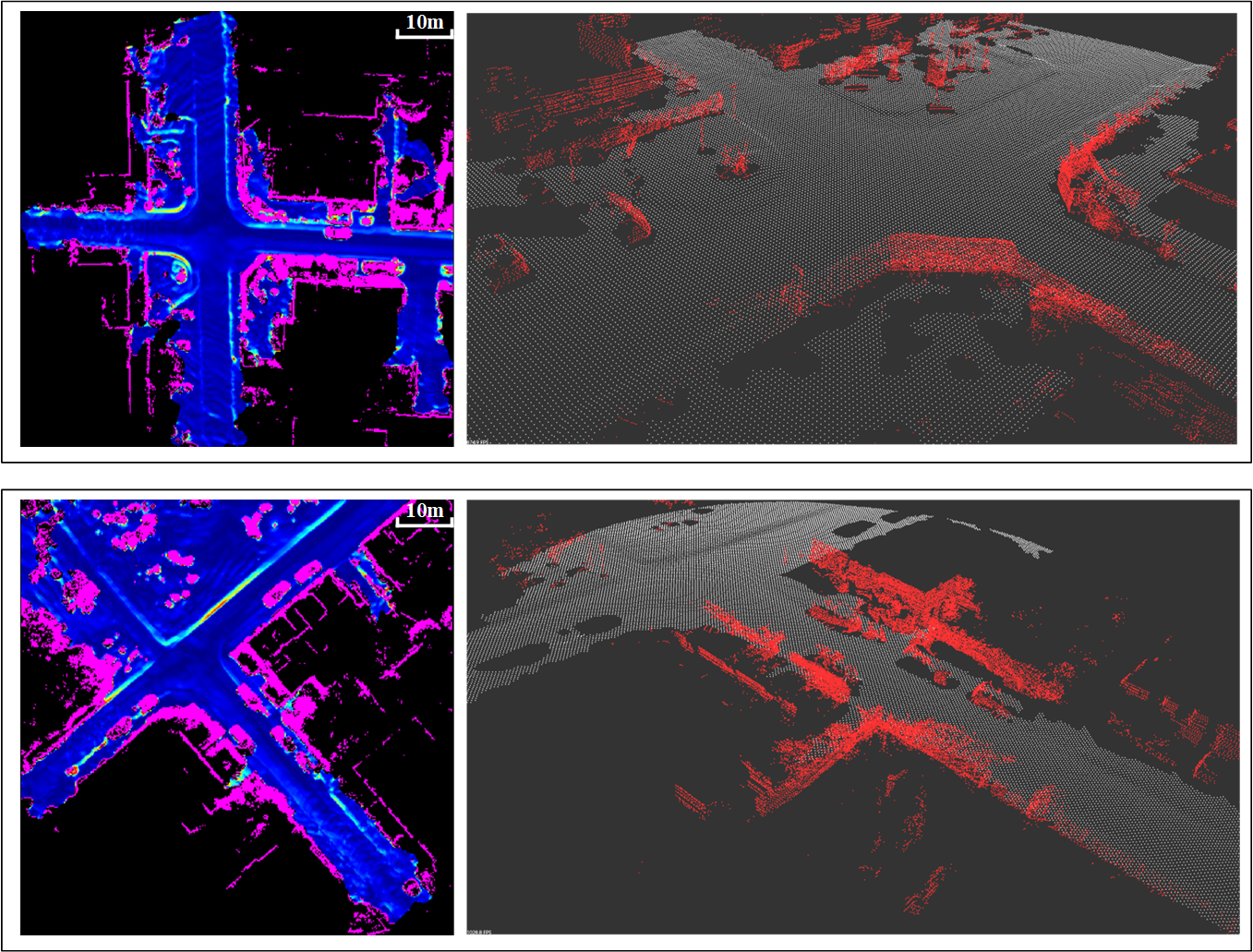}
			\centerline{\footnotesize ({\bf a}) Ours}
			\label{fig_GndNet_a}
		\end{minipage}
	} \vspace{-0.2in} 
	\subfigure{
		\begin{minipage}[b]{0.48\linewidth}
			\centering
			\includegraphics[scale=0.28]{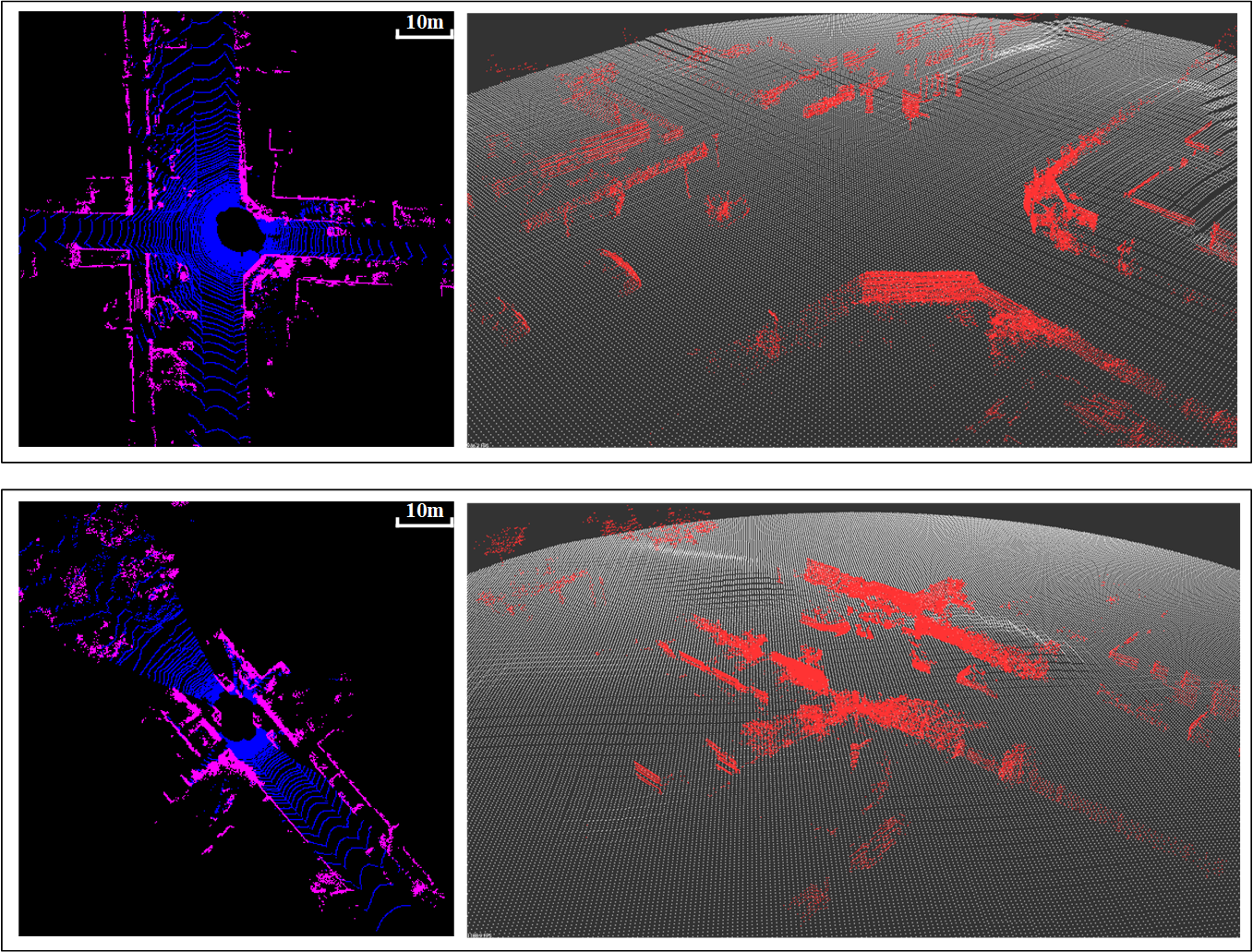}
			\centerline{\footnotesize ({\bf b}) GndNet}
			\label{fig_GndNet_b}
		\end{minipage}
	} 
	\caption{Qualitative comparison results on SemanticKITTI dataset. (a) and (b) are the results generated by the proposed method and the GndNet approach ((the left figures show the traversability maps in 2D, and the right figures show the terrain elevation results in 3D).}
	\label{fig_GndNet}
\end{figure}

\begin{figure}[H]
	\centering 
	\subfigure {
		\begin{minipage}[b]{0.18\linewidth}
			\centering
			\includegraphics[scale=0.32]{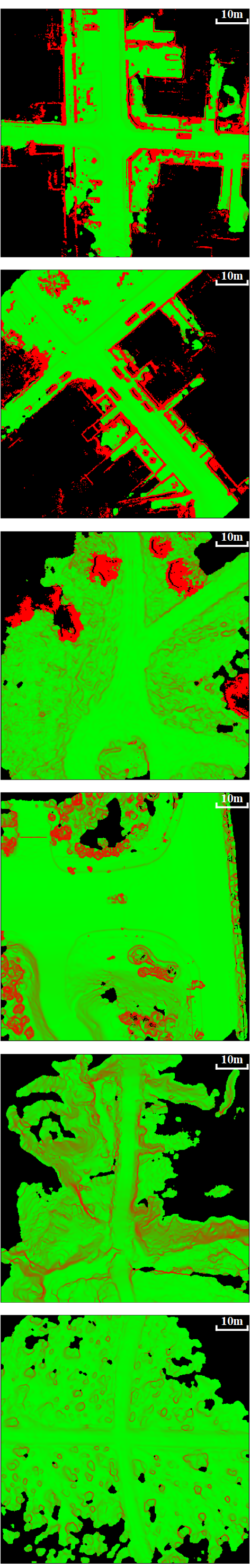}
			\centerline{\footnotesize ({\bf a}) Ground-truth}
			\label{fig_Contrast_normal_GT}
		\end{minipage}
	} %
	\subfigure {
		\begin{minipage}[b]{0.18\linewidth}
			\centering
			\includegraphics[scale=0.32]{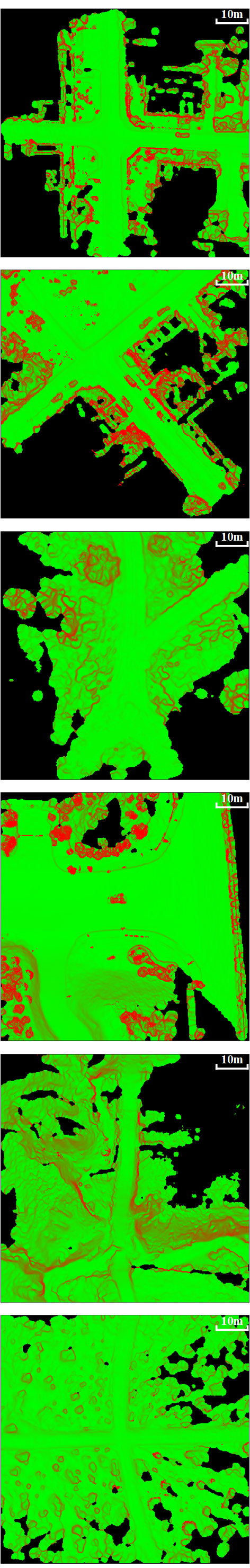}
			\centerline{\footnotesize ({\bf b}) Ours}
			\label{fig_Contrast_normal_our}
		\end{minipage}
	} %
	\subfigure{
		\begin{minipage}[b]{0.18\linewidth}
			\centering
			\includegraphics[scale=0.32]{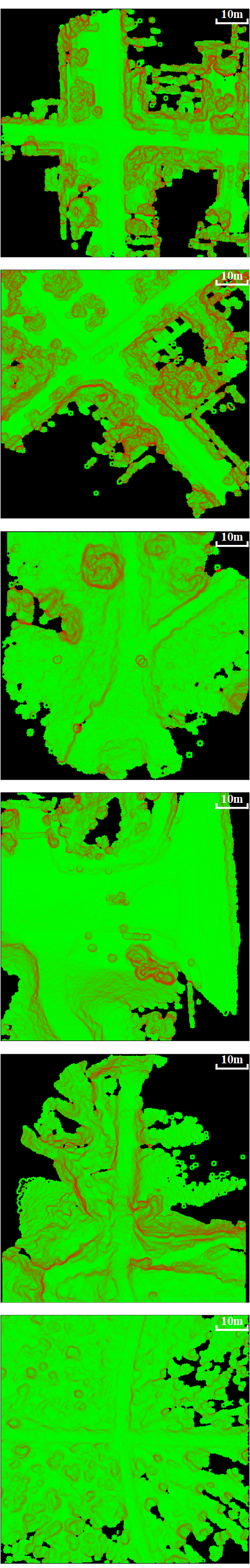}
			\centerline{\footnotesize ({\bf c}) BGK}
			\label{fig_Contrast_normal_BGK}
		\end{minipage}
	} %
	\subfigure{
		\begin{minipage}[b]{0.18\linewidth}
			\centering
			\includegraphics[scale=0.32]{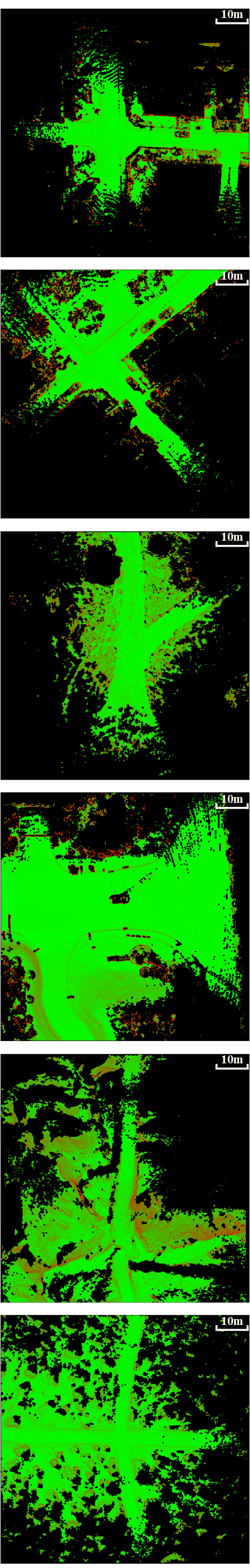}
			\centerline{\footnotesize ({\bf d}) NDT-TM}
			\label{fig_Contrast_normal_NDT}
		\end{minipage}
	} %
	\subfigure{
		\begin{minipage}[b]{0.18\linewidth}
			\centering
			\includegraphics[scale=0.32]{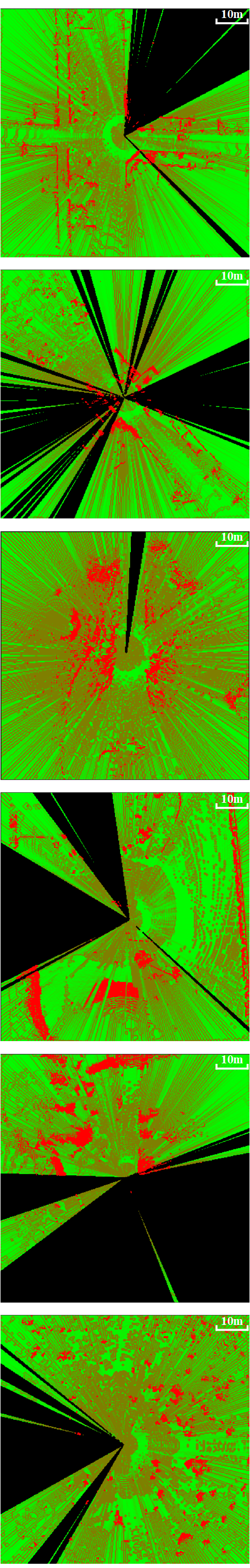}
			\centerline{\footnotesize ({\bf e}) GPR}
			\label{fig_Contrast_normal_GPR}
		\end{minipage}
	} %
	\caption{The illustrative results of normal vector maps generated by different approaches ((a) the ground-truth; (b) the proposed method; (c) the BGK approach \cite{Shan}; (d) the NDT-TM approach \cite{Ahtiainen}; (e) the GPR approach \cite{Chen}). The color of each grid cell encodes the angle between the estimated normal vector and the vertical axis, a higher degree of greenness indicates a more flat terrain.}
	\label{fig_contrast_normal_our}
\end{figure}

\begin{figure}[H]
	\centering
	\subfigure {
		\begin{minipage}[b]{0.18\linewidth}
			\centering
			\includegraphics[scale=0.32]{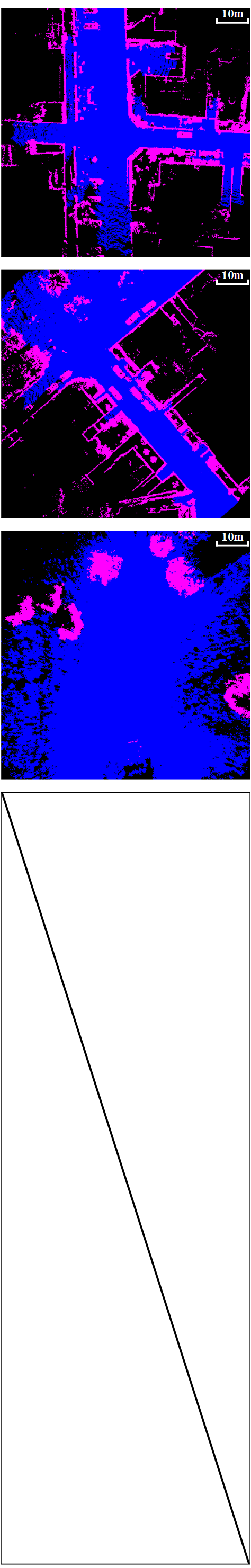}
			\centerline{\footnotesize ({\bf a}) Ground-truth}
			\label{fig_Contrast_label_GT}
		\end{minipage}
	} %
	\subfigure {
		\begin{minipage}[b]{0.18\linewidth}
			\centering
			\includegraphics[scale=0.32]{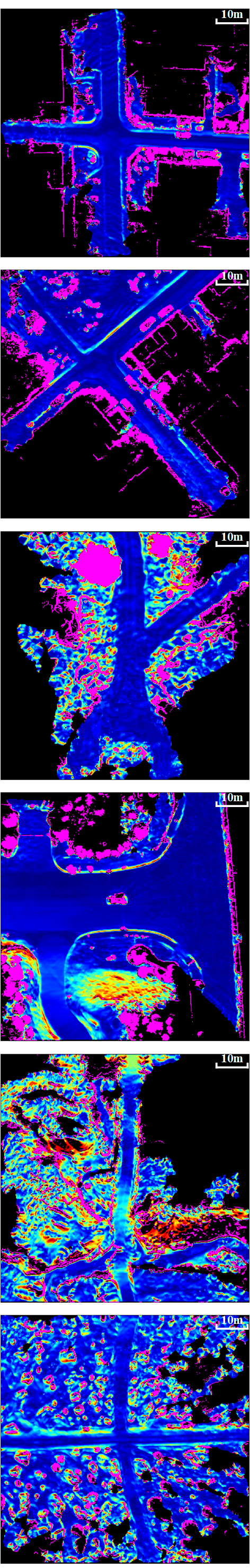}
			\centerline{\footnotesize ({\bf b}) Ours}
			\label{fig_Contrast_label_our}
		\end{minipage}
	} %
	\subfigure{
		\begin{minipage}[b]{0.18\linewidth}
			\centering
			\includegraphics[scale=0.32]{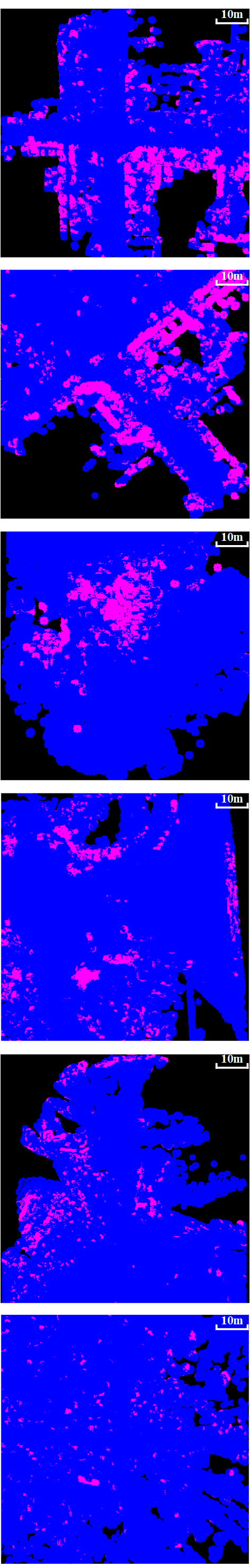}
			\centerline{\footnotesize ({\bf c}) BGK}
			\label{fig_Contrast_label_BGK}
		\end{minipage}
	} %
	\subfigure{
		\begin{minipage}[b]{0.18\linewidth}
			\centering
			\includegraphics[scale=0.32]{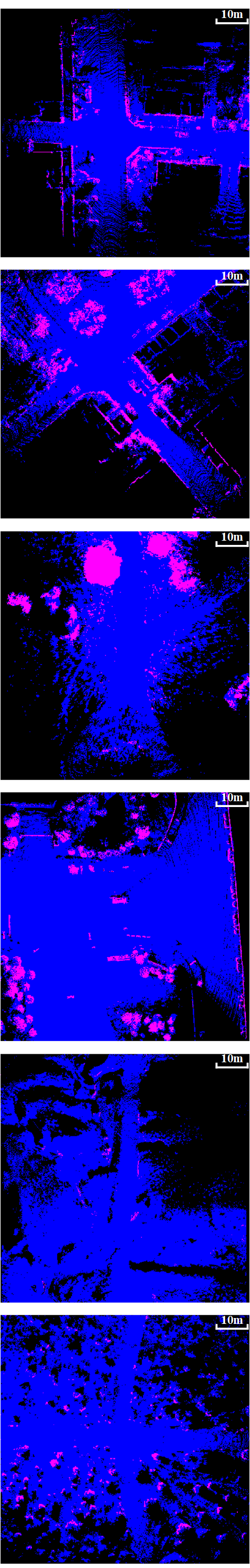}
			\centerline{\footnotesize ({\bf d}) NDT-TM}
			\label{fig_Contrast_label_NDT}
		\end{minipage}
	} %
	\subfigure{
		\begin{minipage}[b]{0.18\linewidth}
			\centering
			\includegraphics[scale=0.32]{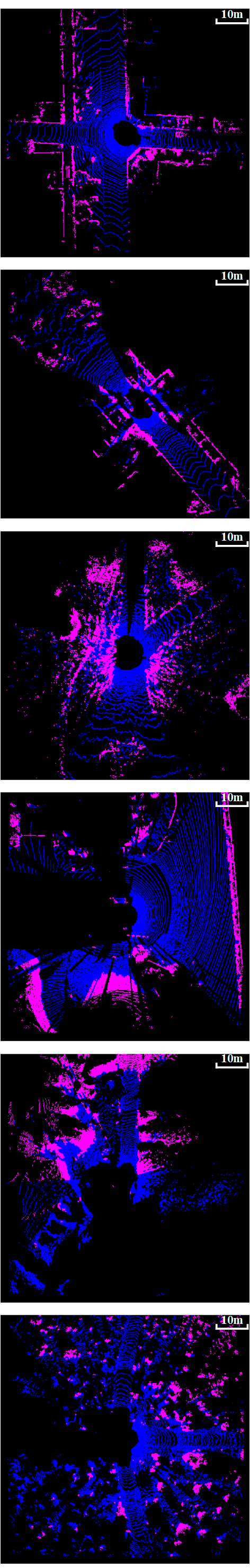}
			\centerline{\footnotesize ({\bf e}) GPR}
			\label{fig_Contrast_label_GPR}
		\end{minipage}
	} %
	\subfigure{
		\begin{minipage}[b]{\linewidth}
			\centering
			\includegraphics[scale=0.325]{figures/Color}
			\label{fig_cost_color}
		\end{minipage}
	} %
	\caption{The illustrative results of traversability maps generated by different approaches ((a) the ground-truth; (b) the proposed method; (c) the BGK approach \cite{Shan}; (d) the NDT-TM approach \cite{Ahtiainen}; (e) the GPR approach \cite{Chen}). Traversable and non-traversable regions are represented by blue and pink, respectively, and (b) is colored by travel cost. The last three rows are results generated in the testing routes of our dataset, so their ground-truth of traversability map is unavailable.}
	\label{fig_contrast_label_our}
\end{figure}

Quantitative results are shown in Table \ref{tabel_Contrast}. Some qualitative comparison results between the proposed method and the GndNet approach on the SemanticKITTI dataset are displayed in Figure \ref{fig_GndNet}, including the 2D traversability maps as well as the elevation estimation results shown in 3D.  Illustrative results of normal vector maps and traversability maps generated by the ground-truth, the proposed method and three non-learning-based traversability analysis approaches are compared in Figure \ref{fig_contrast_normal_our} and Figure \ref{fig_contrast_label_our}.

\begin{table}[H]
	\centering
	\caption{Performance comparison with three state-of-the-art methods.}  
	\begin{tabular}{cccccccc} 
		\toprule
		\textbf{Sequence ID} & \textbf{Approaches} & \tabincell{c}{$\bm{P}$ \\ (\%)} &  \tabincell{c}{$\bm{R}$ \\ (\%)} & \tabincell{c}{$\bm{F_1}$  \\ (\%)} & \tabincell{c}{$\bm{E}$\\ (cm)} & \tabincell{c}{$\bm{R_c}$ \\ (\%)} & \tabincell{c}{\textbf{Time} \\(ms)} \\
		\midrule
		\multirow{5}{*}{S-00} & GPR & 95.44 & 17.76 & 29.95 & 34.40  & 95.71 & 78.32 \\
		& NDT-TM & 91.27 & 67.50 & 77.60 & 4.36 & 65.83 & \textBF{8.50} \\
		& BGK & 86.98 & 74.33 &  80.16 & 10.57 & 77.99 & 135.74 \\
		& GndNet & 96.77 & 17.84 & 30.13 & 39.43 & \textBF{100.00} & 23.61 \\
		& Ours & \textBF{97.72} & \textBF{75.79} & \textBF{85.37} & \textBF{2.37} & 81.83 & 40.10 \\
		\cline{1-8} 
		\multirow{5}{*}{S-05} & GPR & 95.51 & 17.20 & 29.15 & 37.81 & 96.41 & 84.86\\
		& NDT-TM & 91.70 & 68.10 & 78.16 & 5.05 & 64.79 & \textBF{8.81} \\
		& BGK & 87.73 & 75.12 & 80.93 & 10.80 & 77.08 & 138.60 \\
		& GndNet & 96.89 & 17.37 & 29.46 & 39.45 & \textBF{100.00} & 23.07 \\
		& Ours & \textBF{97.08} & \textBF{75.93} & \textBF{85.21} & \textBF{2.27} & 80.22 & 41.30 \\
		\cline{1-8} 
		\multirow{5}{*}{S-07} & GPR & 95.49 & 18.19 & 30.55 & 33.39 & 95.94 & 78.45 \\
		& NDT-TM & 89.78 & 69.17 & 78.14 & 5.01 & 67.94 & \textBF{8.62} \\
		& BGK & 84.67 & 74.29 & 79.14 & 11.86 & 79.27 & 133.22 \\
		& GndNet & 97.32 & 18.35 & 30.88 & 33.30 & \textBF{100.00} & 22.24 \\
		& Ours & \textBF{97.66} & \textBF{77.15} & \textBF{86.20} & \textBF{2.63} & 83.06 & 43.11 \\
		\cline{1-8} 
		\multirow{4}{*}{R-00} & GPR & 95.54 & 21.80 & 35.50 & 55.10 & 93.75 & 81.70 \\
		& NDT-TM & 92.95 & 51.55 & 66.32 & 7.91 & 78.86 & \textBF{9.36} \\
		& BGK & 75.84 & 59.10 & 66.43 & 19.51 & 90.21 & 130.67 \\
		& Ours & \textBF{97.38} & \textBF{62.28} & \textBF{75.97} & \textBF{4.46} & \textBF{94.19} & 48.83 \\
		\cline{1-8} 
		\multirow{4}{*}{R-01} & GPR & 99.58 & 18.67 & 31.44 & 55.02 & \textBF{96.48} & 91.14 \\
		& NDT-TM & 98.67 & 52.65 & 68.66 & 8.58 & 68.39 & \textBF{11.35} \\
		& BGK & 82.10 & 69.85 & 75.48 & 14.39 & 89.57 & 128.23 \\
		& Ours & \textBF{99.69} & \textBF{73.61} & \textBF{84.69} & \textBF{7.28} & 92.70 & 47.80 \\
		\cline{1-8} 
		\multirow{4}{*}{R-02} & GPR & 99.13 & 24.88 & 39.78 & 65.83 & 93.71 & 89.07 \\
		& NDT-TM & 97.91 & 58.67 & 73.37 & 5.91 & 79.47 & \textBF{9.82} \\
		& BGK & 85.07 & 70.26 & 76.96 & 11.34 & 91.28 & 124.16 \\
		& Ours & \textBF{99.89} & \textBF{72.59} & \textBF{84.08} & \textBF{4.27} & \textBF{94.50} & 52.38  \\
		\cline{1-8} 
		\multirow{4}{*}{O-00} & GPR &  \multicolumn{3}{c}{\multirow{4}*{\diagbox[width=8em]{\,}{\,}}} & 41.32 & 65.05 & 131.88  \\
		& NDT-TM &  &  &  & 2.95 & 82.34 & \textBF{10.77}  \\
		& BGK &   &  &  & 4.94 & 93.52 & 147.02 \\
		& Ours &  &  &  & \textBF{1.45} & \textBF{96.56} & 65.41 \\
		\cline{1-8} 
		\multirow{4}{*}{O-01} & GPR &  \multicolumn{3}{c}{\multirow{4}*{\diagbox[width=8em]{\,}{\,}}} & 108.12 & 59.81 & 105.48 \\
		& NDT-TM &  &  &  &  5.75 & 77.90 & \textBF{10.98} \\
		& BGK &  &  &  & 8.54 & 86.98 & 144.47 \\
		& Ours &  &  &  & \textBF{4.27} & \textBF{91.97} & 66.77 \\
		\cline{1-8}  
		\multirow{4}{*}{O-02} & GPR &  \multicolumn{3}{c}{\multirow{4}*{\diagbox[width=8em]{\,}{\,}}} & 89.29 & 62.44 & 145.73 \\
		& NDT-TM &  &  &  & 4.68 & 76.15 & \textBF{10.83} \\
		& BGK &  &  &  &  5.81 & 89.89 & 153.04 \\
		& Ours &  &  &  & \textBF{3.56} & \textBF{94.66} & 74.40  \\
		\bottomrule	
	\end{tabular}
	\label{tabel_Contrast}
\end{table}               

$\mathbf{Discussion}$: 

(1) Compared with the other four methods, the proposed method achieves the best results on most performance metrics ($P$, $R$, $F_1$, and $E$), which can generate complete terrain models and accurate traversability analysis results in real-time. In addition, as shown in Figure \ref{fig_Contrast_label_our}, regions with high travel costs (such as road curbs, ditches, slopes, and road boundaries) can be clearly distinguished, which can provide the path planning module with richer information;

(2) The GndNet approach \cite{Paigwar} has the highest performance on $R_c$ and a good performance on $P$, but a poor performances on $R$ and $E$. It can also be observed from Figure \ref{fig_GndNet_b} that its elevation estimation results has a low accuracy since the road curbs can not be distinguished. In contrast, the road curb can be clearly observed from Figure \ref{fig_GndNet_a}. The reason of highest $R_c$ is that the end-to-end network can predict the elevation of any regions around the UGV, and the reason of low $R$ score is that this approach only analyzes the traversability of regions containing observed points in the current LiDAR frame;

(3) Although the original BGK approach \cite{Shan} achieves relatively good completeness performance ($R$ and $R_c$), its accuracy performances ($P$ and $E$) is poor and it has the highest computational complexity. The reason of low accuracy is that this method exploits the range differences between adjacent laser scans to detect obstacles, which has poor results in complex outdoor environments. In addition, this method also uses the BGK inference to estimate the traversability of each grid cell. This reasoning process makes the terrains near the obstacles are easily estimated as non-traversable regions, so the obstacles will be much larger than their true size in the traversability analysis results (as shown in Figure \ref{fig_Contrast_label_BGK}). It is worth mentioning that the performance metric $E$ is improved in the testing routes of our dataset since there are fewer obstacles in our testing environments;

(4) The NDT-TM method \cite{Ahtiainen} has the lowest computational complexity, and can obtain good performance on terrain elevation estimation ($E$). However, this method cannot predict the elevation, resulting in a low completeness score (as shown in Figure \ref{fig_Contrast_normal_NDT} and \ref{fig_Contrast_label_NDT});  

(5) The GPR method \cite{Chen} achieves relatively good performance on $R_c$ and $P$ on SemanticKITTI and RELLIS-3D dataset, but poor performance on $R$ and $E$. The reason of low $R$ score is that this method only analyzes the traversability of regions containing observed points in the current LiDAR frame. Besides, as this method simplifies 2D GPR into a series of 1D GPRs, the resulting terrain models are discontinuous (as shown in Figure \ref{fig_Contrast_normal_GPR}) and have low $E$ score. In addition, performance metrics $E$ and $R_c$ of this method drop in our testing routes since its poor adaptability to the off-road environments and the limited horizontal FOV of LiDAR.   
\begin{figure}[H]
	\centering
	\includegraphics[scale=0.49]{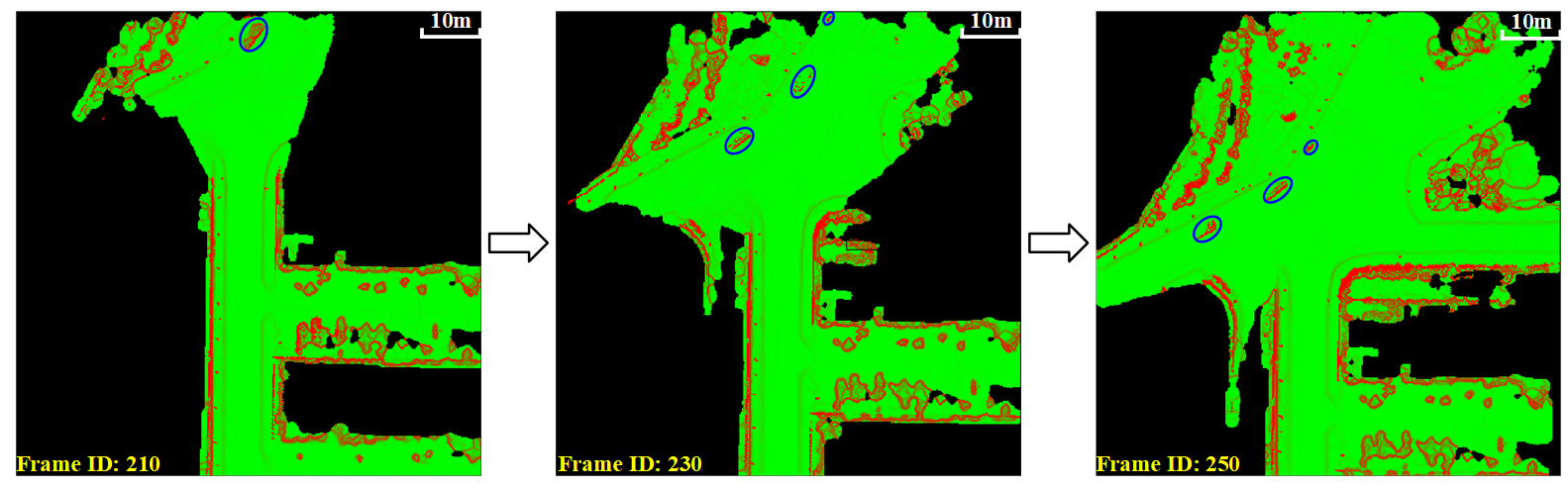} \\
	\caption{Qualitative results of normal vector maps in dynamic environments. The blue ellipsoids enclose two moving vehicles and a moving bicyclist.}
	\label{fig_Dynamic}
\end{figure}

\subsection{Experiments in Dynamic Environments}
We also evaluate the proposed method in a dynamic environment. In this environment, two vehicles and a bicyclist are moving from opposite directions (shown as the blue ellipsoids in Figure \ref{fig_Dynamic}). The results of normal vector maps in consecutive frames are shown in Figure \ref{fig_Dynamic}.

$\mathbf{Discussion}$: Dynamic objects do have a negative impact on approaches processing consecutive frames. But in our approach, thanks to the specifically designed variance filter, the utilization of bilateral filter and the local convexity analysis, our method is quite robust to those dynamic objects. From Figure \ref{fig_Dynamic}, it is observed that the proposed method could output stable results across consecutive frames, and the dynamic objects do not influence our terrain modeling result.

\begin{figure}[t]
	\centering
	\subfigure{
		\begin{minipage}[b]{\linewidth}
			\centering
			\includegraphics[scale=0.4]{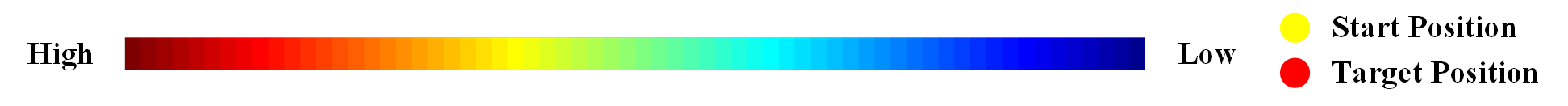}
			\label{fig_plan_up}
		\end{minipage}
	} %
	\subfigure {
		\begin{minipage}[b]{0.23\linewidth}
			\centering
			\includegraphics[scale=0.38]{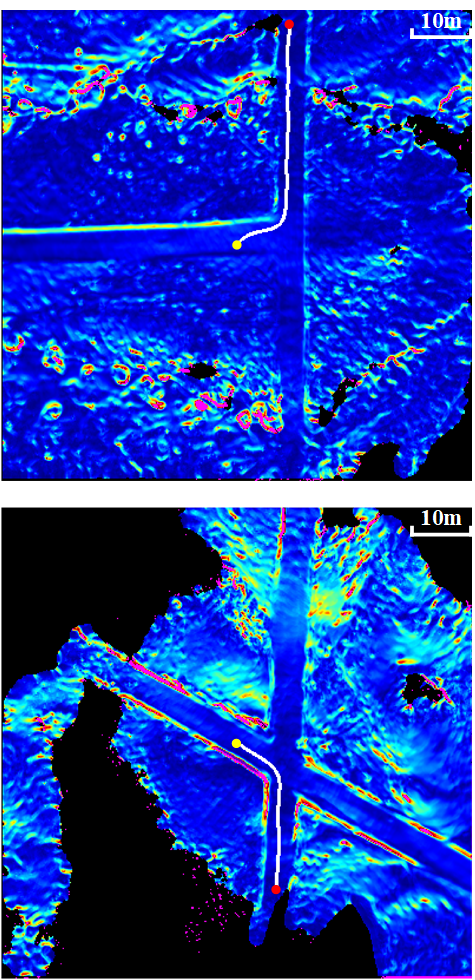}
			\centerline{\footnotesize ({\bf a}) Ours}
			\label{fig_path_our}
		\end{minipage}
	} %
	\subfigure{
		\begin{minipage}[b]{0.23\linewidth}
			\centering
			\includegraphics[scale=0.38]{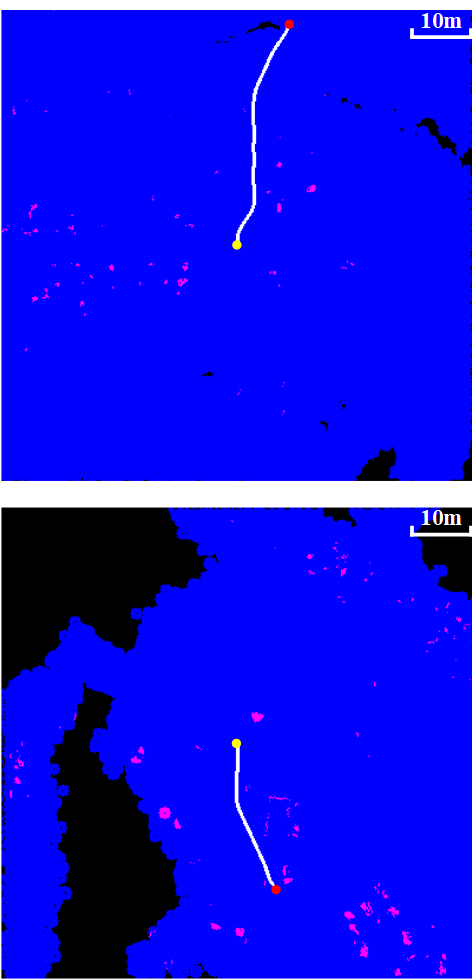}
			\centerline{\footnotesize ({\bf b}) BGK}
			\label{fig_path_BGK}
		\end{minipage}
	} %
	\subfigure{
		\begin{minipage}[b]{0.23\linewidth}
			\centering
			\includegraphics[scale=0.38]{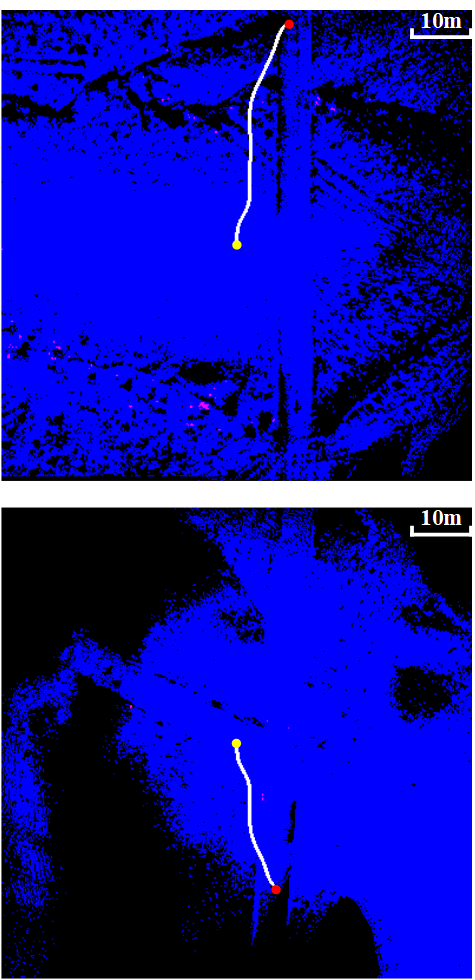}
			\centerline{\footnotesize ({\bf c}) NDT-TM}
			\label{fig_path_NDT}
		\end{minipage}
	} %
	\subfigure{
		\begin{minipage}[b]{0.23\linewidth}
			\centering
			\includegraphics[scale=0.38]{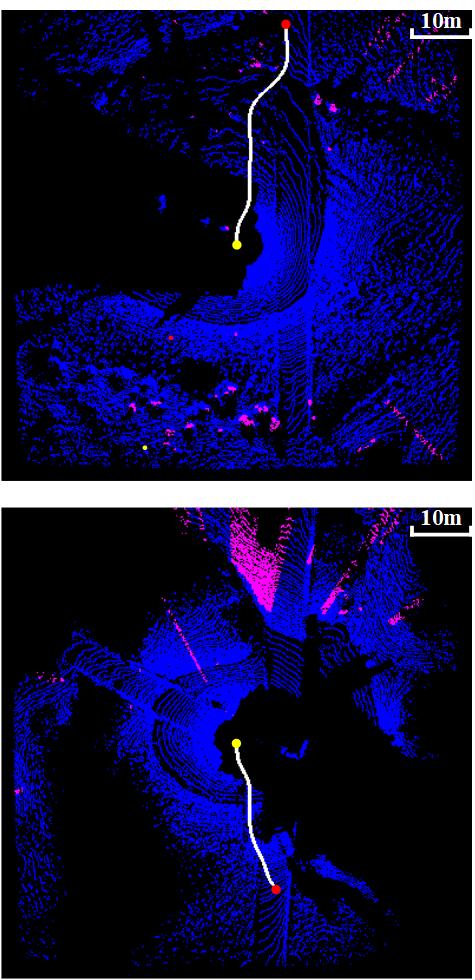}
			\centerline{\footnotesize ({\bf d}) GPR}
			\label{fig_path_GPR}
		\end{minipage}
	} %
	\caption{The optimal paths (the white curves) from the start positions (the yellow circles) to the target positions (the red circles) generated by the same path planning algorithm from different cost maps (The cost maps are generated by (a) the proposed method; (b) the BGK approach \cite{Shan}; (c) the NDT-TM approach \cite{Ahtiainen}; (d) the GPR approach \cite{Chen}, respectively.) Traversable and non-traversable regions are represented by blue and pink, respectively. (a) is colored by travel cost.}
	\label{fig_path}
\end{figure}
\subsection{Path Planning Experiments}
\label{Section: Path_Planning}
The cost map generated by the proposed method could also aid the path planning module. Given the cost maps generated by the proposed method and the other three traversability analysis approaches (the BGK approach \cite{Shan}, the NDT-TM approach \cite{Ahtiainen}, and the GPR approach \cite{Chen}), an improved hybrid A* algorithm \cite{A_star} is applied to find an optimal path from a start position to a target position. The travel cost of each grid cell is calculated by Equation (\ref{equation:travel_cost}) in the proposed method, which is a continuous value. For the cost maps generated by the other three approaches, the travel cost of each grid is merely a binary discrete value (represent traversable or non-traversable). The optimal paths generated on different cost maps are compared in Figure \ref{fig_path}\label{key}. The white curves represent the calculated optimal paths. The red and yellow circles indicate the start positions and the target positions.

$\mathbf{Discussion}$: It can be seen that the planned paths from our cost map are always located on the road, while the paths generated by the other three approaches deviate to the outside of the road. The reason is that the travel costs of different traversable regions can be well distinguished in the cost maps generated by the proposed method, but the travel costs of different traversable regions are same in the cost maps generated by the other three methods. This experiment proves that the proposed method can assist UGVs to select reasonable and safe paths in complex environments.

\section{CONCLUDING REMARKS}
\label{Section_Conclusion}

In this paper, a novel LiDAR-based traversability analysis approach has been proposed. It can generate stable, complete and accurate terrain models and traversability analysis results in real-time. Information from consecutive frames are fused in a NDT mapping approach. By utilizing the spatial-temporal Bayesian generalized kernel inference and bilateral filtering, the terrain elevation of those unobserved regions could be efficiently inferred from their neighboring observed regions, and the completeness of the generated terrain models will be improved. Then, terrain traversability is estimated by geometric connectivity analysis between neighboring terrain cells, and terrains with different traversability costs can be discriminated. Experimental results both on publicly available dataset and our own dataset show that the proposed method outperforms state-of-the-art approaches.

Our technical contributions are summarized as follows:

The first contribution is that we make full use of the information provided from consecutive LiDAR frames for traversability analysis, rather than treating it as a single-frame task. A NDT mapping approach is utilized to model the terrain. In addition, we also consider the quantization error between the global grid map and local grid map, and use the residual to eliminate the quantization error. By adopting this multi-frame fusion approach, some estimation errors can be easily avoided, and the estimation results are more likely to be stable and complete.

The second contribution is that we propose a spatial-temporal BGK elevation inference method. Compared to the original BGK approach, we make two improvements. The first one is that we introduce bilateral filtering to the BGK elevation inference, thus alleviating the edge blurring problem. The second improvement is that the elevation variance estimated by NDT mapping approach is introduced as a weight in the BGK inference. With the help of this weight, grid cells with larger variance will contribute less to elevation inference. By applying these two improvements, the estimated terrain models and traversability analysis results can tend to be more accurate.

Thirdly, by analyzing the geometric connectivity properties between adjacent terrain cells, we could obtain a travel cost map. This cost map is helpful for distinguishing different terrain types, such as road curbs, ditches, slopes, and road boundaries. Therefore, the proposed method can assist the UGV path planning module to select reasonable and safe paths in complex environments.

However, the proposed method still has some limitations. One problem is that the proposed method can not deal with negative obstacles that exists in the environment. The negative obstacles are observed as some empty regions in the terrain map, and our method tends to over-smooth these empty regions by the BGK inference. This problem will be addressed in our future works. In addition, the computational efficiency and the completeness of the detection results can be further improved by utilizing a multi-resolution framework, this is also left as a future work.

\subsection*{\textcolor{black}{Appendix}}
\subsubsection*{Proof 1: The transformation of $p\left(\bm{\theta}_*^t|\bm{x}_*, \bm{O}^t\right)$}
This subsection shows the whole transformation process of $p\left(\bm{\theta}_*^t|\bm{x}_*, \bm{O}^t\right)$, which is also the derivation of Equation (\ref{equation:Bayesian_4}).
   
Firstly, $p\left(\bm{\theta}_*^t|\bm{x}_*, \bm{O}^t\right)$ can be expanded by Bayes' theorem and conditional independence assumption:
\begin{equation}
\begin{split}
p\left(\bm{\theta}_*^t|\bm{x}_*, \bm{O}^t\right) &= \int_{\bm{\theta}_{1:N_o^t}^t} p\left(\bm{\theta}_{1:N_o^t}^t, \bm{\theta}^t_* | \bm{x}_*, \bm{x}_{1:N_o^t}, \mathcal{N}_{1:N_o^t}^t\right) d\bm{\theta}^t_{1:N_o^t} \\
&= \int_{\bm{\theta}_{1:N_o^t}^t} \frac{p\left(\mathcal{N}_{1:N_o^t}^t | \bm{\theta}_{1:N_o^t}^t, \bm{\theta}^t_*, \bm{x}_*, \bm{x}_{1:N_o^t} \right) \cdot p\left(\bm{\theta}_{1:N_o^t}^t, \bm{\theta}^t_* | \bm{x}_*, \bm{x}_{1:N_o^t}\right)}{p\left(\mathcal{N}_{1:N_o^t}^t | \bm{x}_*, \bm{x}_{1:N_o^t}\right)} d\bm{\theta}^t_{1:N_o^t} \\
&= \int_{\bm{\theta}_{1:N_o^t}^t} \frac{p\left(\mathcal{N}_{1:N_o^t}^t | \bm{\theta}_{1:N_o^t}^t\right) \cdot p\left(\bm{\theta}_{1:N_o^t}^t, \bm{\theta}^t_* | \bm{x}_*, \bm{x}_{1:N_o^t}\right)}{p\left(\mathcal{N}_{1:N_o^t}^t | \bm{x}_{1:N_o^t}\right)} d\bm{\theta}^t_{1:N_o^t} \\
& \propto \int_{\bm{\theta}_{1:N_o^t}^t} p\left(\mathcal{N}_{1:N_o^t}^t | \bm{\theta}_{1:N_o^t}^t\right) \cdot p\left(\bm{\theta}_{1:N_o^t}^t, \bm{\theta}^t_* | \bm{x}_*, \bm{x}_{1:N_o^t}\right) d\bm{\theta}^t_{1:N_o^t} 
\end{split}
\label{equation:Bayesian_Bayes}
\end{equation}
where $\bm{\theta}^t_{1:N_o^t}$ represents the latent parameters associated with the potential terrain cells $\bm{x}_{1:N_o^t}$ at time step $t$. According to Law of Total Probability and conditional independence assumption, $p\left(\mathcal{N}_{1:N_o^t}^t | \bm{\theta}_{1:N_o^t}^t\right)$ in Equation (\ref{equation:Bayesian_Bayes}) can be expanded as:
\begin{equation}
p\left(\mathcal{N}_{1:N_o^t}^t | \bm{\theta}_{1:N_o^t}^t\right) = \prod_{i=1}^{N_o^t} p\left(\mathcal{N}_i^t|\bm{\theta}_i^t \right) \,.
\label{equation:Bayesian_Total_Probability}
\end{equation}

Substitute Equation (\ref{equation:Bayesian_Total_Probability}) into Equation (\ref{equation:Bayesian_Bayes}), $p\left(\bm{\theta}_*^t|\bm{x}_*, \bm{O}^t\right)$ can be further expressed as: 
\begin{equation}
p\left(\bm{\theta}_*^t|\bm{x}_*, \bm{O}^t\right) \propto \int_{\bm{\theta}_{1:N_o^t}^t} \prod_{i=1}^{N_o^t} p\left(\mathcal{N}_i^t|\bm{\theta}_i^t\right) \cdot p\left(\bm{\theta}^t_{1:N_o^t}, \bm{\theta}^t_*|\bm{x}_*, \bm{x}_{1:N_o^t}\right) d\bm{\theta}^t_{1:N_o^t} \,.
\label{equation:Bayesian_Bayes2}
\end{equation}

Any parameter $\bm{\theta}^t_i$ in set $\bm{O}^t$ are conditionally independent during the BGK inference, thus $p\left(\bm{\theta}^t_{1:N_o^t}, \bm{\theta}^t_*|\bm{x}_*, \bm{x}_{1:N_o^t}\right)$ in Equation (\ref{equation:Bayesian_Bayes2}) can also be expanded by Law of Total Probability and conditional independence assumption:
\begin{equation}
\begin{split}
p\left(\bm{\theta}^t_{1:N_o^t}, \bm{\theta}^t_*|\bm{x}_*, \bm{x}_{1:N_o^t}\right) &= p\left(\bm{\theta}^t_{1:N_o^t} | \bm{\theta}^t_*, \bm{x}_*, \bm{x}_{1:N_o^t}\right) \cdot p\left(\bm{\theta}^t_* |  \bm{x}_*, \bm{x}_{1:N_o^t}\right)  \\
&= \prod_{i=1}^{N_o^t} p\left(\bm{\theta}^t_{i} |  \bm{\theta}^t_*, \bm{x}_*, \bm{x}_{i}\right) \cdot p\left(\bm{\theta}^t_* |  \bm{x}_*\right) \,.
\end{split}
\label{equation:Bayesian_Total_Probability2}
\end{equation} 

Substitute Equation (\ref{equation:Bayesian_Total_Probability2}) into Equation (\ref{equation:Bayesian_Bayes2}) and marginalize the latent parameters, $p\left(\bm{\theta}_*^t|\bm{x}_*, \bm{O}^t\right)$ can be simplified as:
\begin{equation}
\begin{split}
p\left(\bm{\theta}_*^t|\bm{x}_*, \bm{O}^t\right) &\propto \int_{\bm{\theta}^t_{1:N_o^t}} \prod_{i=1}^{N_o^t} p\left(\mathcal{N}_i^t|\bm{\theta}^t_i\right) \cdot p\left(\bm{\theta}^t_{i} | \bm{\theta}_*^t, \bm{x}_*, \bm{x}_{i}\right) \cdot p\left(\bm{\theta}_*^t | \bm{x}_*\right) d\bm{\theta}^t_{1:N_o^t}  \\
& \propto \prod_{i=1}^{N_o^t} p\left(\mathcal{N}_i^t | \bm{\theta}_*^t, \bm{x}_*, \bm{x}_{i}\right) \cdot p\left(\bm{\theta}_*^t | \bm{x}_*\right) \,.
\end{split}
\label{equation:Bayesian_3}
\end{equation}

In the smooth extended likelihood model \cite{Brown}, the distribution $ g = p\left(\mathcal{N}_i^t | \bm{\theta}_*^t, \bm{x}_*, \bm{x}_{i}\right)$ is called the extended likelihood distribution, and the distribution $f = p\left(\mathcal{N}_i^t | \bm{\theta}_*^t\right)$ is called the likelihood distribution. The Kullback-Leibler divergence $D_{\mathrm{KL}}\left(g||f\right)$ of the likelihood distribution $f$ from the extended likelihood distribution $g$ is bounded by the function $\rho: \mathcal{X} \times \mathcal{X} \rightarrow  \mathbb{R}^{+}$, which can be denoted as $D_{\mathrm{KL}}\left(g||f\right) \leq \rho\left(\bm{x}_{i}, \bm{x}_*\right)$, and the following theorem has been proved in \cite{Brown}:

\begin{theorem}
	The maximum entropy distribution $g$ satisfying $D_{\mathrm{KL}}\left(g||f\right) \leq \rho\left(\bm{x}_{i}, \bm{x}_*\right)$ has the form $g \propto f^{k\left(\bm{x}_{i}, \bm{x}_*\right)}$, where $k: \mathcal{X} \times \mathcal{X} \rightarrow \left[0, 1\right]$ is a kernel function which can be uniquely determined from $\rho\left(\cdot,\cdot\right)$ and $f\left(\cdot,\cdot\right)$.
\end{theorem}
According to Theorem 1, Equation (\ref{equation:Bayesian_3}) can be further transformed to: 
\begin{equation}
p\left(\bm{\theta}_*^t|\bm{x}_*, \bm{O}^t\right) \propto \prod_{i=1}^{N_o^t} p\left(\mathcal{N}_i^t | \bm{\theta}_*^t\right)^{k\left(\bm{x}_{i}, \bm{x}_*\right)} \cdot p\left(\bm{\theta}_*^t | \bm{x}_*\right) \,.
\end{equation}

\subsubsection*{Proof 2: The relationship between $p\left(\bm{\theta}_*^t|\bm{x}_*, \bm{O}^t\right)$ and $p\left(\mathcal{N}_*^t|\bm{x}_*,\bm{O}^t\right)$}
The purpose of the spatial-temporal BGK inference approach is to estimate the elevation $h_*$ of the target grid cell $\bm{x}_*$, and the distribution to be estimated for this regression problem should be $p\left(\mathcal{N}_*^t|\bm{x}_*,\bm{O}^t\right)$ in Equation (\ref{equation:Bayesian_1}). This distribution can be further inferred by $p\left(\bm{\theta}_*^t|\bm{x}_*, \bm{O}^t\right)$ and $p\left(\mathcal{N}_*^t|\bm{\theta}_*^t\right)$. For the distribution $p\left(\bm{\theta}_*^t|\bm{x}_*, \bm{O}^t\right)$, it can be expressed as a normal distribution $\mathcal{N}\left(\mu_*^t, \Sigma_*^t\right)$. The distribution $p\left(\mathcal{N}_*^t|\bm{\theta}_*^t\right)$ also satisfies a normal distribution $\mathcal{N}\left(\tilde{\mu}_*^t, \hat{\Sigma}_*^t\right)$ since we have assumed that the likelihood distribution $p\left(\mathcal{N}_i^t|\bm{\theta}_*^t\right)$ satisfies $\mathcal{N}\left(\tilde{\mu}_i^t, \hat{\Sigma}_i^t\right)$ when transforming Equation (\ref{equation:Bayesian_4}) to Equation (\ref{equation:Bayesian_5}). Expressing $p\left(\bm{\theta}_*^t|\bm{x}_*, \bm{O}^t\right)$ and $p\left(\mathcal{N}_*^t|\bm{\theta}_*^t\right)$ in the form of normal distributions $\mathcal{N}\left(\mu_*^t, \Sigma_*^t\right)$ and $\mathcal{N}\left(\tilde{\mu}_*^t, \hat{\Sigma}_*^t\right)$, Equation (\ref{equation:Bayesian_1}) can be transformed as:
\begin{equation}
\begin{split}
&p\left(\mathcal{N}_*^t|\bm{x}_*,\bm{O}^t\right) = \int \frac{1}{\sqrt{2\pi \hat{\Sigma}_*^t}} \exp\left[-\frac{\left(h_* - \tilde{\mu}_*^t\right)^2}{2\hat{\Sigma}_*^t}\right] \cdot \frac{1}{\sqrt{2\pi \Sigma_*^t}} \exp\left[-\frac{\left(\tilde{\mu}_*^t - \mu_*^t\right)^2}{2\Sigma_*^t}\right] d\tilde{\mu}_*^t \\
& \propto  \int \exp \left[-\frac{\left(\Sigma_*^t+\hat{\Sigma}_*^t\right) {\tilde{\mu}_*^t}{}^2 - 2\left(h_*\Sigma_*^t + \mu_*^t\hat{\Sigma}_*^t\right)\tilde{\mu}_*^t + \left({h_*}^2\Sigma_*^t + {\mu_*^t}^2\hat{\Sigma}_*^t\right)}{2\hat{\Sigma}_*^t \Sigma_*^t} \right] d\tilde{\mu}_*^t \\
& \propto \exp \left[ -\frac{\left({h_*}^2\Sigma_*^t + {\mu_*^t}^2\hat{\Sigma}_*^t\right)}{2\hat{\Sigma}_*^t \Sigma_*^t}\right] \int \exp \left[-\frac{{\tilde{\mu}_*^t}{}^2 -2\frac{h_*\Sigma_*^t + \mu_*^t\hat{\Sigma}_*^t}{\Sigma_*^t+\hat{\Sigma}_*^t}\tilde{\mu}_*^t + \left(\frac{h_*\Sigma_*^t + \mu_*^t\hat{\Sigma}_*^t}{\Sigma_*^t+\hat{\Sigma}_*^t}\right)^2 - \left(\frac{h_*\Sigma_*^t + \mu_*^t\hat{\Sigma}_*^t}{\Sigma_*^t+\hat{\Sigma}_*^t}\right)^2}{\frac{2\hat{\Sigma}_*^t \Sigma_*^t}{\Sigma_*^t+\hat{\Sigma}_*^t}} \right] d\tilde{\mu}_*^t \\
& \propto \exp \left[ -\frac{\left({h_*}^2\Sigma_*^t + {\mu_*^t}^2\hat{\Sigma}_*^t\right)}{2\hat{\Sigma}_*^t \Sigma_*^t} + \frac{\left(\frac{h_*\Sigma_*^t + \mu_*^t\hat{\Sigma}_*^t}{\Sigma_*^t+\hat{\Sigma}_*^t}\right)^2}{\frac{2\hat{\Sigma}_*^t \Sigma_*^t}{\Sigma_*^t+\hat{\Sigma}_*^t}}\right] \int  \exp \left[-\frac{\left(\tilde{\mu}_*^t - \frac{h_*\Sigma_*^t + \mu_*^t\hat{\Sigma}_*^t}{\Sigma_*^t+\hat{\Sigma}_*^t}\right)^2}{\frac{2\hat{\Sigma}_*^t \Sigma_*^t}{\Sigma_*^t+\hat{\Sigma}_*^t}} \right] d\tilde{\mu}_*^t \\
& \propto \exp \left[ -\frac{\left({h_*}^2\Sigma_*^t + {\mu_*^t}^2\hat{\Sigma}_*^t\right) \left(\Sigma_*^t+\hat{\Sigma}_*^t\right)}{2\hat{\Sigma}_*^t \Sigma_*^t \left(\Sigma_*^t+\hat{\Sigma}_*^t\right)} + \frac{\left({h_*}^2{\Sigma_*^t}^2 + 2h_*\Sigma_*^t \mu_*^t \hat{\Sigma}_*^t + {\mu_*^t}^2{\hat{\Sigma}_*^t}{}^2\right)}{2\hat{\Sigma}_*^t \Sigma_*^t \left(\Sigma_*^t+\hat{\Sigma}_*^t\right)} \right] \\
& \propto \exp \left[ -\frac{\Sigma_*^t \hat{\Sigma}_*^t {h_*}^2 - 2\Sigma_*^t \mu_*^t \hat{\Sigma}_*^t h_* + {\mu_*^t}^2\hat{\Sigma}_*^t \Sigma_*^t}{2\hat{\Sigma}_*^t \Sigma_*^t \left(\Sigma_*^t+\hat{\Sigma}_*^t\right)} \right] \\ 
& \propto \exp \left[-\frac{\left(h_* - \mu_*^t\right)^2}{2\left(\Sigma_*^t + \hat{\Sigma}_*^t\right)}\right].\\
\end{split}
\label{equation:Bayesian_8}
\end{equation} 
It can be proved that distributions $p\left(\mathcal{N}_*^t|\bm{x}_*,\bm{O}^t\right)$ and $p\left(\bm{\theta}_*^t|\bm{x}_*, \bm{O}^t\right)$ have the same mean $\mu_*^t$.

\subsubsection*{Acknowledgments}
This work was supported by the National Natural Science Foundation of China under No. 61790565 and No. 61803380.

\bibliographystyle{apalike}
\bibliography{jfrExampleRefs}

\end{document}